\documentclass[lettersize,journal]{IEEEtran}
\usepackage{amsmath, amsfonts}
\usepackage[linesnumbered,ruled]{algorithm2e}
\usepackage{algorithmic}
\usepackage{graphicx}
\usepackage{textcomp}
\usepackage{xcolor}
\usepackage{algorithm2e}
\usepackage{multirow}
\usepackage[normalem]{ulem}
\usepackage{caption}
\usepackage{graphicx}
\usepackage{float} 
\usepackage{enumitem}
\usepackage{subcaption}
\usepackage{diagbox}
\usepackage{times}
\usepackage{array}
\usepackage{tabularray}
\usepackage{marvosym}
\begin{document}

\title{An End-to-End Model for Time Series Classification In the Presence of Missing Values}

\author{Pengshuai Yao, Mengna Liu, Xu Cheng, Fan Shi, Huan Li, Xiufeng Liu, Shengyong Chen
\thanks{Pengshuai Yao, Xu Cheng, Fan Shi, and Shengyong Chen are with the School of Computer Science and Engineering, Tianjin University of Technology, Tianjin, China. Xiufeng Liu is with the Department of Technology, Management and Economics, Technical University of Denmark, Produktionstorvet, Denmark, 2800. Huan Li is with the Department of Computer Science, Aalborg University, Aalborg, Denmark,  9220. }}

\markboth{}%
{Shell \MakeLowercase{\textit{et al.}}: A Sample Article Using IEEEtran.cls for IEEE Journals}


\maketitle

\begin{abstract}
Time series classification with missing data is a prevalent issue in time series analysis, as temporal data often contain missing values in practical applications. The traditional two-stage approach, which handles imputation and classification separately, can result in sub-optimal performance as label information is not utilized in the imputation process. On the other hand, a one-stage approach can learn features under missing information, but feature representation is limited as imputed errors are propagated in the classification process. To overcome these challenges, this study proposes an end-to-end neural network that unifies data imputation and representation learning within a single framework, allowing the imputation process to take advantage of label information. Differing from previous methods, our approach places less emphasis on the accuracy of imputation data and instead prioritizes classification performance. A specifically designed multi-scale feature learning module is implemented to extract useful information from the noise-imputation data. The proposed model is evaluated on 68 univariate time series datasets from the UCR archive, as well as a multivariate time series dataset with various missing data ratios and 4 real-world datasets with missing information. The results indicate that the proposed model outperforms state-of-the-art approaches for incomplete time series classification, particularly in scenarios with high levels of missing data. 
\end{abstract}


\section{Introduction}

\IEEEPARstart{T}{ime} series classification is a crucial area of study within the realm of machine learning \cite{ma_ajrnn, li2022ips}. In recent years, a plethora of models have been proposed for time series classification \cite{2020InceptionTime, onmi_tang}. Despite the progress made in this field, a common assumption among these methods is that the time series data used are complete and without missing values \cite{de2019gru, cheng2021novel}. However, in practical applications, it is not uncommon for time series data to be affected by missing values of varying frequencies, due to various factors such as anomalies \cite{WANG2022823}, communication errors, or malfunctioning sensors \cite{li2018missing}.

\begin{figure}[htbp]
	\centering
			\includegraphics[width=\linewidth]{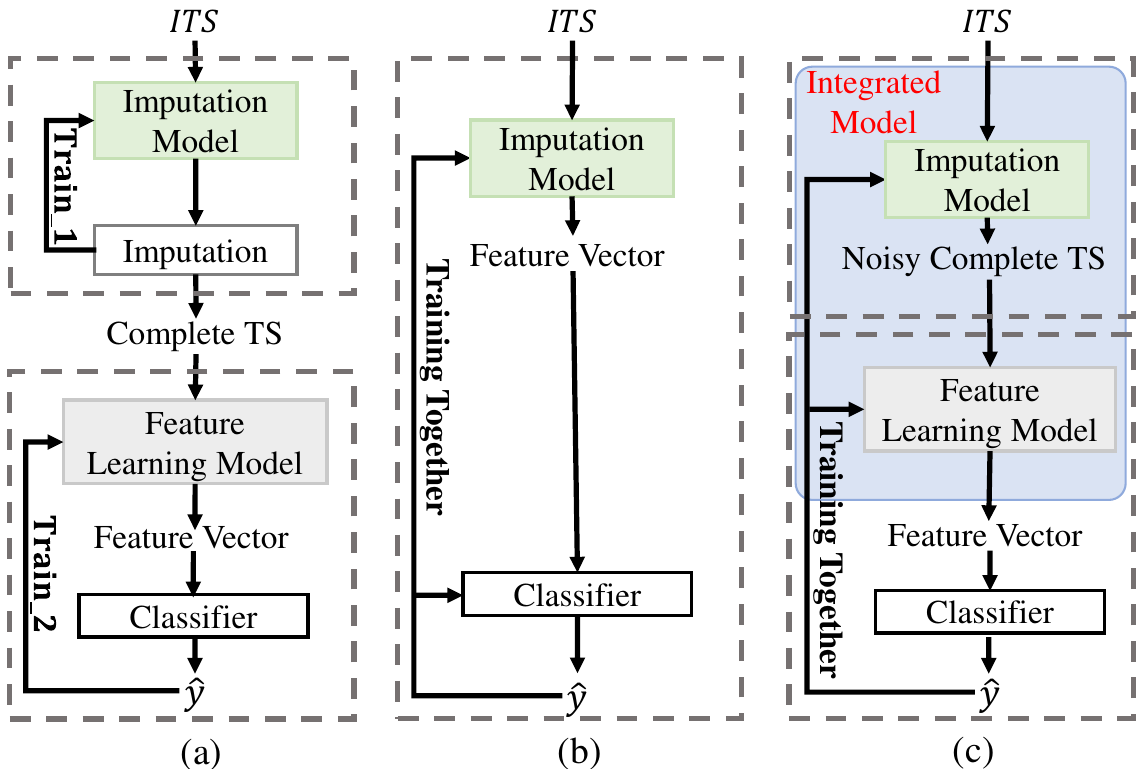}
	\caption{(a) two-stage method and (b) one-stage method, that utilizes features from the imputation network for classification, while our proposed method in (c) considers the output of the imputation model to be noisy, thus unifies data imputation and feature learning within the same framework. }
	\label{COMPARE}
  \vspace{-4mm}
\end{figure}

The presence of missing values in the data can impede the inference process and compromise the results of data analysis \cite{du2022saits}. The task of classifying time series data with missing values, referred to as incomplete time series classification (ITSC), presents a more practical and challenging problem than traditional time series classification with complete data. ITSC requires the ability to identify patterns and dynamics within time series data in the presence of incomplete information. Consequently, addressing the negative impact of missing values in ITSC models is a crucial concern within the field of time series data mining. This is especially true in cases where missing data are missing completely at random, as this can introduce bias and lead to inaccurate results. Therefore, it is necessary to develop methods to address ITSC problems in a robust and effective manner.

Imputation is an indispensable step in ITSC, ensuring the availability of data. However, concerning downstream classification tasks, current research predominantly focuses on improving the credibility of Imputation data. For data without Ground-Truth, existing methods often employ metrics such as Mean Squared Error (MSE) or Root Mean Squared Error (RMSE) to calculate observable values, thereby assessing the overall Imputation capability. However, it's important to note that these metrics may not be adequate for determining the usability of Imputation values. The conventional approach for handling ITSC is a two-stage process \cite{zhang2021missing, sefidian2019missing}, as depicted in Fig.~\ref{COMPARE} (a), which first estimates missing values and then performs the classification. However, this approach has some limitations. First, the imputation is typically treated as a separate process of pre-processing data, without interacting with classifier training, which can lead to suboptimal results \cite{wells2013strategies, che2018recurrent}. Second, classification performance is highly dependent on the chosen imputation method, as biased interpolated values can greatly impair classification results. Therefore, it is crucial to identify the cause of the missing data and use the appropriate method. Last, the two-stage approaches do not utilize label information well during the imputation process, while label information always plays a crucial role in improving model performance.

\begin{figure}
     \centering
     \begin{subfigure}[b]{0.225\textwidth}
         \centering
         \includegraphics[width=\textwidth]{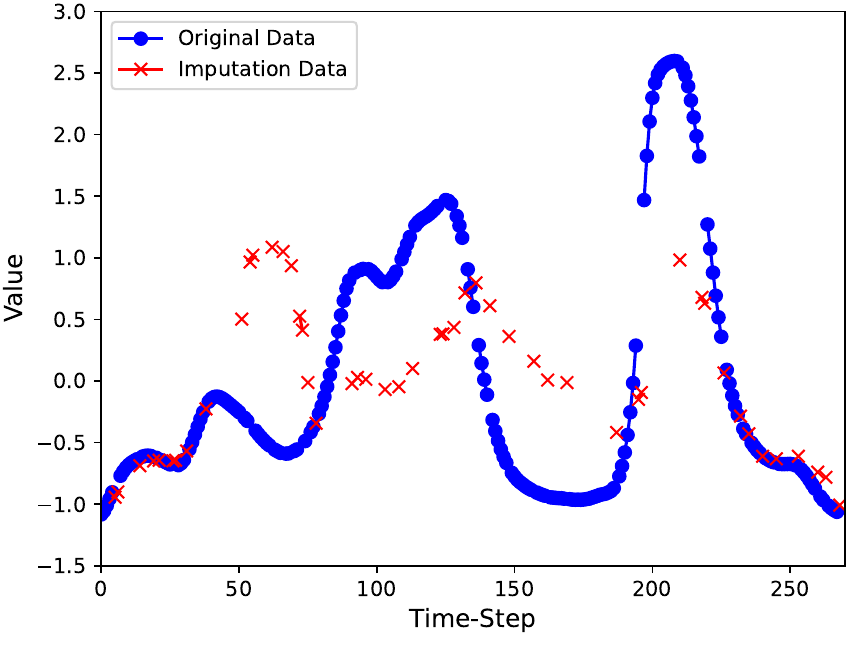}
         \caption{20\%}
     \end{subfigure}
     \hfill
     \begin{subfigure}[b]{0.225\textwidth}
         \centering
    \includegraphics[width=\textwidth]{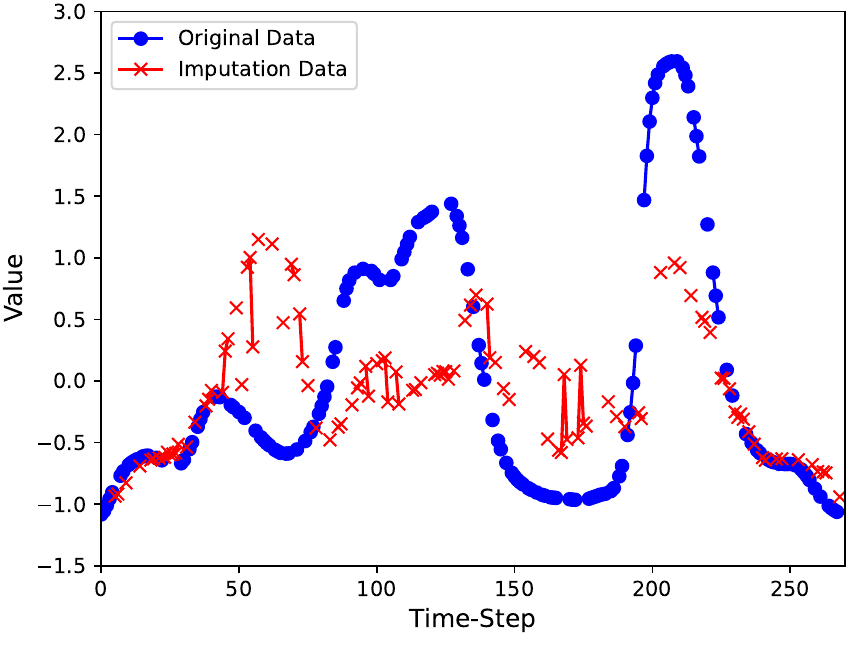}
         \caption{40\%}
     \end{subfigure}

     \begin{subfigure}[b]{0.225\textwidth}
         \centering
         \includegraphics[width=\textwidth]{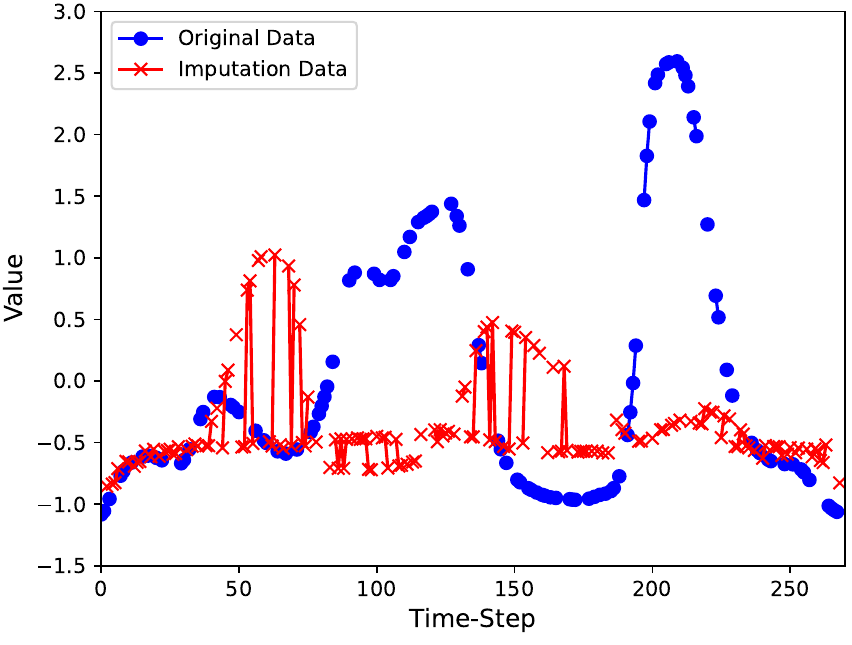}
         \caption{60\%}
     \end{subfigure}
     \hfill
     \begin{subfigure}[b]{0.225\textwidth}
         \centering
         \includegraphics[width=\textwidth]{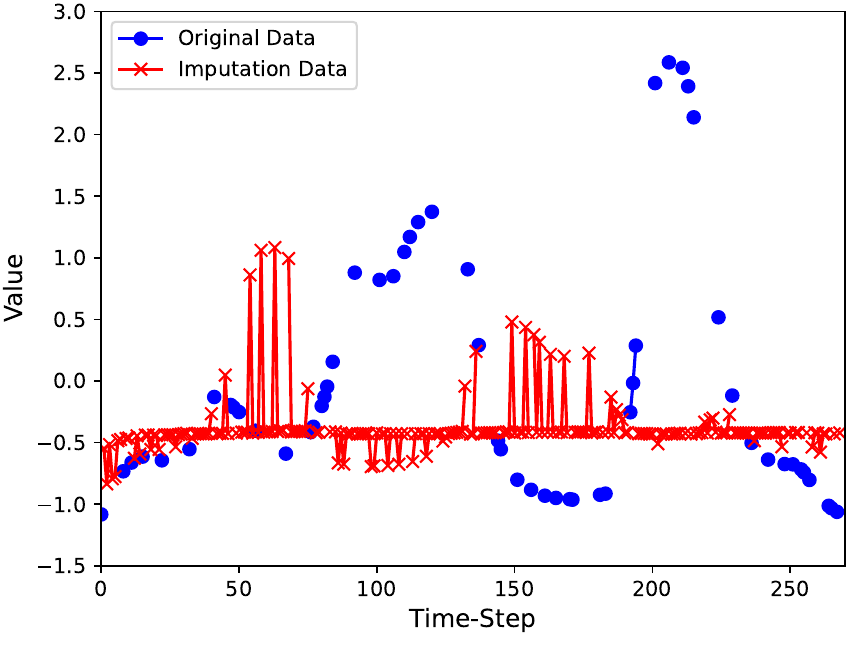}
         \caption{80\%}
     \end{subfigure}     
        \caption{Imputation results of BRITS in different missing ratios.}
        \label{fig:Imp}
     \vspace{-3mm}
\end{figure}

To address the above limitations, several studies \cite{che2018recurrent,weerakody2021review} have proposed the joint learning approach, which combines missing value imputation with classification during training. There are two main approaches in joint learning: (1) modifying traditional temporal learning models, such as Recurrent Neural Networks (RNNs), to handle missing data and (2) combining RNN with a Generative Adversarial Network (GAN) for missing data imputation. For the first approach, the study \cite{che2018recurrent} shows that a carefully designed missing temporal learning module can significantly improve the classification performance. For the second approach, the study \cite{weerakody2021review}
demonstrates that combining RNN with GAN can effectively learn temporal information and impute missing data in time series, thanks to the powerful learning capabilities of GANs. The study shows that this method can still achieve good learning performance from incomplete time series data. Despite these promising results, joint learning-based approaches for ITSC still face the following major challenges:

\begin{itemize}[leftmargin=*]
\item \textbf{Imputation difficulty}:  A common strategy is to employ model-based approaches to learn missing patterns and observable values for imputing missing data. As shown in Fig.~\ref{fig:Imp}, the effectiveness of the BRITS \cite{cao2018brits} method is demonstrated under various missing rates. In the case of completely random missing data, as the missing rate increases, the imputation performance of this method significantly deteriorates. Moreover, after imputation, it completely alters the original temporal data trends and peak values, thereby adversely affecting downstream classification tasks. Additionally, Limited data volume, resulting in unreliable or even infeasible Imputation. The distinctive nature of time series data introduces a significant error bias in time-series classification, particularly when addressing anomalies during interpolation.

\item \textbf{Feature representation}: The above methods rely on their unique structures to extract hidden features in missing time series data, with the assumption that utilizing these features to construct a classifier will lead to desirable outcomes. However, the features generated directly from the imputation model (as demonstrated in Fig.~\ref{COMPARE} (b)) can have limited representation as imputation errors may not be effectively eliminated and could accumulate to some degree within the generated features.


\end{itemize}

To tackle the aforementioned challenges, this paper presents a novel joint learning model with superior feature representational capability for ITSC. Our model acknowledges that imputed missing information may contain errors, even when the training accuracy is high. Hence, our model focuses on learning \emph{useful} information from ``imputation noise" data. To achieve this, we first use an imputation model to perform an initial imputation on the missing data, followed by a specially designed feature learner to extract hidden information from the ``imputation noise" data (as depicted in Fig.~\ref{COMPARE} (c)). By jointly training the model for data imputation and feature learning, we allow the label information to be leveraged by the imputation model, mitigating the impact of imputation errors on performance.

The contributions of this work are as follows:
\begin{enumerate}[leftmargin=*]
    \item A novel deep joint learning model is proposed for ITSC that combines data imputation and feature learning. This model prioritizes feature representation over data imputation, providing a unified framework that leverages label information to improve the imputation process. This distinguishes it from prior work that mainly focuses on missing data imputation.
    
    \item The proposed model employs a simple Gated Recurrent Unit (GRU) model to impute missing data, not overly emphasizing the credibility of imputation values, followed by a multi-scale large kernel model associated with dilation convolution. The model incorporates both a multi-scale mechanism and a large kernel size to capture various-range correlations, reducing the negative effects of noisy imputed values generated by the GRU as well as reducing training difficulty.

    \item Extensive experiments have been conducted, including comparisons with four state-of-the-art methods, using 68 univariate time series datasets from the UCR archive with varying missing ratios, and four real-world univariate time series datasets. Additionally, the model is evaluated on a multivariate time series dataset under two missing scenarios and five missing ratios. The results demonstrate the superiority of the proposed model, and sensitivity and ablation experiments further validate the effectiveness of its design.
\end{enumerate}

In this paper, we assume the missing completely at random (MCAR) regime, as did in \cite{ma_ajrnn}. MCAR refers to missingness that is independent of all observed and unobserved values. One such example is sensor failure, as in the UCR $DodgerLoopDay$ dataset.

The remainder of this paper is structured as follows: Section \ref{related} reviews the progress of missing information learning for time series data. Section \ref{method} presents the model structure, details of each component, and the training process. Section \ref{experiment} describes  the experimental. Section \ref{conclusion} concludes the paper.

\section{Related Work}\label{related}
This section will review the related work on ITSC, including both two-stage and one-stage approaches.

\subsection{Two-stage Approach for ITSC}

The two-stage approach for incomplete time series classification starts with estimating the missing values, followed by applying a classification method to the resulting complete data. 

Various methods have been proposed to handle missing data in time series classification. Simple methods like removing observations with missing values or replacing them with fixed values, such as zero, mean, or nearest observed value, can result in the loss of important information or time-dependent relationships \cite{cao2018brits}. Advanced methods rooted in statistics and machine learning have been proposed, including the K-nearest neighbor (KNN) approach \cite{cheng2019novel}, the Expectation Maximization (EM) algorithm \cite{josse2019consistency}, autoregressive (AR) based algorithms \cite{9380704}, and matrix decomposition algorithms \cite{9380704}. The KNN method is based on individual samples and does not consider temporal dependencies, while the EM algorithm is constrained by linearity assumptions. The AR-based method estimates missing data using previous values but involves a complex training process. Matrix decomposition methods rely on low-rank assumptions to make the imputation.

Several deep learning-based methods, such as RNN, GANs, and Transformer-based models, have been proposed for time series imputation \cite{yoon2018gain, luo2019e2gan, li2019misgan, luo2018multivariate}. These methods leverage deep learning's ability to capture complex patterns and relationships in time series data. GANs, in particular, have gained popularity for imputing missing values in time series.
Yoon et al. proposed GAIN \cite{yoon2018gain}, a GAN-based method for imputing missing values in time series. Li et al. proposed MisGAN \cite{li2019misgan}, which uses a GAN structure to perform imputation. Luo et al. introduced GRUI \cite{luo2018multivariate}, a GAN structure constructed by RNN, and E$^{2}$GAN \cite{luo2019e2gan}, an GRUI-based autoencoder, for irregular time series imputation. In addition, Liu et al. presented NAOMI, a non-autoregressive model reinforced by adversarial training \cite{NEURIPS2019_50c1f44e}. Shan et al. presented NRTSI, a imputation method based on a Transformer encoder \cite{2021NRTSI}. Du et al. proposed SAITS, a state-of-the-art multi-headed self-attention Transformer model with a diagonal mask for time series imputation \cite{du2022saits}. 

While deep learning-based imputation methods have shown remarkable performance, they solely focus on handling missing data and do not optimize the classification process, resulting in subpar outcomes \cite{wells2013strategies, che2018recurrent}. Hence, there is a necessity for additional research to devise a holistic approach that effectively addresses missing values while optimizing time series classification.

\subsection{One-stage Approach for ITSC}

The one-stage approach for ITSC combines the imputation and classification tasks within a single model, instead of treating them as separate stages. This approach has been shown to offer advantages over the traditional two-stage method by optimizing both tasks simultaneously and providing an end-to-end framework for modeling incomplete time series data. Che et al. introduced GRU-D \cite{che2018recurrent}, a variation of the gated recurrent unit (GRU) that employs time decay of the last observed value to impute missing data. M-RNN \cite{yoon2018estimating} and BRITS \cite{cao2018brits} also use RNNs to infer missing values, but M-RNN considers them as constants while BRITS treats them as variables in the RNN graph. Moreover, BRITS accounts for feature correlation, which is not considered in M-RNN.
Ma et al. proposed AJRNN \cite{ma_ajrnn}, which uses a discriminator to supervise the imputation task performed by GRU. Under the Missing Completely At Random setting, they proposed an RSU-based one-stage approach \cite{shen2018end} for handling missing data in time series classification. The approach involves training a network to predict the observation using an exponential decay of past hidden unit activations (RSU). A GAN network is used to supervise the imputation task in a joint training approach \cite{ma_ajrnn}, however, this approach has limitations as it only considers the RNN-generated features for classification and can be challenging to train. Zhang et al. proposed Raindrop \cite{zhang2021graph}, a graph-guided network for irregularly sampled time series. RAINDROP learns a distinct sensor dependency graph for every sample capturing time-varying dependencies between sensors.

The above one-stage methods have made great strides in addressing missing temporal data, but they all learn the data based on their unique structures and output the required features for classification directly through imputation networks. However, these features may not be sufficient to fully express the underlying data characteristics due to the presence of imputation errors. In this paper, we focus on the task of time series classification in the presence of missing information, with a greater emphasis on the accuracy of classification rather than the quality of missing data imputation. To this end, we propose a framework that unifies imputation and classification by jointly training these tasks and treating the complete data generated by the imputation model as noisy input. Our approach learns category-related information from this noisy data to perform classification.

\section{Method} \label{method}

\begin{figure*}[t!]
  \centering
  \includegraphics[width=\linewidth]{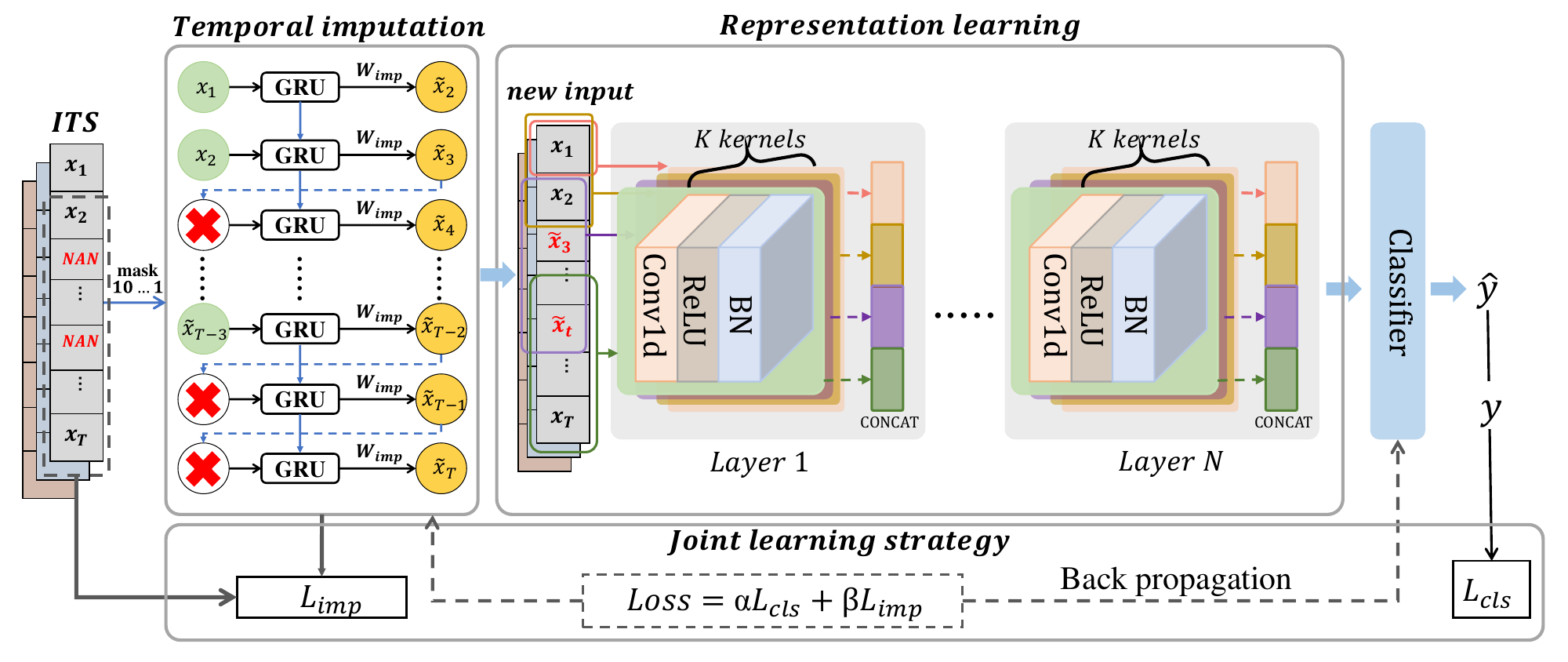}
  \caption{Our ITSC network framework, taking a univariate sample as an example, the input is ITS, passing through a temporal imputation module to impute missing data, using a GRU structure combined with observable values. For the noisy input after imputation, we use a multi-scale feature learning module, which includes N layers of multi-scale 1D CNNs, and finally, classification is performed and optimized through joint learning.}
  \label{Model}
  \vspace{-3pt}
\end{figure*}

\subsection{Problem Formulation}

\textbf{Definition 1} (Time Series). A $T$-step complete time series $\boldsymbol{I}=\left\{\boldsymbol{i}_{1}, \boldsymbol{i}_{2}, \ldots, \boldsymbol{i}_{T}\right\}$, where $\boldsymbol{i}_{t} \in \mathbb{R}^{n}$ and $n$ is dimension size, $t$ is the timestamp.

\textbf{Definition 2} (Incomplete Time Series). A $T$-step incomplete time series $\boldsymbol{X}=\left\{\boldsymbol{x}_{1}, \boldsymbol{x}_{2}, \ldots, \boldsymbol{x}_{T}\right\}$, where $\boldsymbol{x}_{t} \in \mathbb{R}^{n}$ and $n$ is dimension size, $t$ is the timestamp. And, when the $t$ timestamp is missing, the value of $\boldsymbol{x}_{t}$ is $nan$.

\textbf{Definition 3} (Imputation For Incomplete Time Series). For incomplete time series, temporal imputation aims to use simple GRU model timestamp-by-timestamp processing, and when this timestamp $t$ is missing, the hidden state $\boldsymbol{h}_{t-1}$ of the GRU is used to estimate the current missing value $\boldsymbol{x}_{t}$ up to $\tilde{\boldsymbol{x}}_{t}$. Afterwards, we take $\boldsymbol{x}_{t}$, represented by the timestamps that are not missing in the incomplete time series, and $\tilde{\boldsymbol{x}}_{t}$, estimated by the missing timestamps, and obtain our temporary complete time series by mask vector $\boldsymbol{m}_{t} \in \left\{0,1\right\}$, and $\tilde{\boldsymbol{x}}_{t}$ is also iteratively updated by the learning of GRU model during the network training process.

\textbf{Definition 4} (Representation Learning For Incomplete Time Series Classification). Incomplete time series are first obtained by temporary imputation modeling to obtain temporary complete time series. Given a collection of incomplete time series after temporal imputation $\mathrm{X} =\left\{\boldsymbol{X}_{1}, \boldsymbol{X}_{2}, \ldots, \boldsymbol{X}_{N}\right\}$ of N instances and its corresponding label set  $\mathrm{Y} =\left\{{y}_{1}, {y}_{2}, \ldots, {y}_{N}\right\}$, representation learning for classification aims to obtain category-specific representations $\mathrm{Z} =\left\{{Z}_{1}, {Z}_{2}, \ldots, {Z}_{N}\right\}$ by learing a  multi-scale convolutional neural networks. The representation vector ${Z}_{i} =\left\{\boldsymbol{z}_{i,1}, \boldsymbol{z}_{i,2}, \ldots, \boldsymbol{z}_{i,T}\right\}$ contains representation vectors $\boldsymbol{z}_{i,t} \in \mathbb{R}^{k}$ for each timestamp $t$, where $k$ is the dimension of representation vectors and $T$ is the length of the time series. 

\subsection{Overview}

Fig.~\ref{Model} presents an overview of the proposed model, which is composed of three main components: a temporal imputation module (TIM), a multi-scale feature learning module (MSFL), and a joint learning strategy.
They will be detailed from Section~\ref{ssec:temporal_imputation} to Section~\ref{ssec:joint_learning}, respectively.

To address the missing information in the incomplete time series, we adopt the widely used GRU model \cite{chung2014empirical, cho2014learning} to estimate and impute missing values. The GRU cells encode the time series at each time step, and the hidden state is then fed into a fully-connected layer to obtain an estimated value. During the imputation process, the estimated value from the previous time step is used if the current time step is missing, otherwise, the original data is preserved. In this manner, the incomplete time series is transformed into a complete series by combining both imputed and original values.

To address the potential errors in the imputed data, a multi-scale feature learning module is implemented. This is crucial as missing values have a random nature, and it is essential to design a feature learner that can effectively extract relevant information. To achieve this, we employ a multi-scale Convolutional Neural Network (CNN) with a large kernel size. This design allows for the effective extraction of useful information from the imputed input. The module consists of $N$ layers with multi-large-kernel 1D CNNs that use dilation convolution to capture various range dependencies in the input as well as reduce the training difficulty.

In the joint learning process, the imputation and classification tasks are trained simultaneously, allowing label information to guide both imputation and feature learning. The imputation loss $\mathcal{L}_{\mathbf{imp}}$ is calculated as the mean squared error between the estimated and true values, and the classification loss $\mathcal{L}_{\mathbf{c l s}}$ is calculated as the softmax cross-entropy loss between the predicted and ground-truth labels. The final loss is a linear combination of these two losses, and the model is trained using back-propagation. This approach leverages the benefits of joint learning and robust feature extraction, thereby reducing the negative impact of missing values on the classification performance.

\subsection{Temporal Imputation} \label{ssec:temporal_imputation}

Temporal imputation is the first module in the proposed model, which can be formalized as follows. Given a continuous $T$-step time series $\boldsymbol{X}=\left\{\boldsymbol{x}_{1}, \boldsymbol{x}_{2}, \ldots, \boldsymbol{x}_{T}\right\}$, where $\boldsymbol{x}_{t} \in \mathbb{R}^{n}$ and $n$ is dimension size, the encoding of $\boldsymbol{X}$ through a GRU yields a sequence of hidden features $\boldsymbol{H}=\left\{\boldsymbol{h}_{1}, \boldsymbol{h}_{2}, \ldots, \boldsymbol{h}_{T}\right\}$, where the hidden state $\boldsymbol{h}_{t} \in \mathbb{R}^{m}$ and $m$ is the dimension of hidden size. The operation of a GRU can be described by the following formulas:
\begin{equation}
   \boldsymbol{z}_{t}=\sigma\left(\boldsymbol{W}_{x z} \boldsymbol{x}_{t}+\boldsymbol{W}_{h z} \boldsymbol{h}_{t-1}+\boldsymbol{b}_{z}\right),
   \label{EQ1}
\end{equation}
\begin{equation}
   \boldsymbol{r}_{t}=\sigma\left(\boldsymbol{W}_{x r} \boldsymbol{x}_{t}+\boldsymbol{W}_{h r} \boldsymbol{h}_{t-1}+\boldsymbol{b}_{r}\right),
   \label{EQ2}
\end{equation}
\begin{equation}
   \tilde{\boldsymbol{h}}_{t}=\tanh \left(\boldsymbol{W}_{x h} \boldsymbol{x}_{t}+\boldsymbol{W}_{h h}\left(\boldsymbol{r}_{t} \odot \boldsymbol{h}_{t-1}\right)+ \boldsymbol{b}_{h}\right),
   \label{EQ3}
\end{equation}
\begin{equation}
   \boldsymbol{h}_{t}=\boldsymbol{z}_{t} \odot \tilde{\boldsymbol{h}}_{t}+\left(1-\boldsymbol{z}_{t}\right) \odot \boldsymbol{h}_{t-1},
   \label{EQ4}
\end{equation}
where the sigmoid function, $\sigma(\cdot)$, is used to calculate the update gate $\boldsymbol{z}$ and reset gate $\boldsymbol{r}$. The parameters for the update gate, $\boldsymbol{W}_{x z}$ and $\boldsymbol{W}_{h z}$, and the parameters for the reset gate, $\boldsymbol{W}_{x r}$ and $\boldsymbol{W}_{h r}$, are used in the calculation. The internal state $\tilde{\boldsymbol{h}}_{t}$ is calculated using the parameters $\boldsymbol{W}_{x h}$ and $\boldsymbol{W}_{h h}$. The element-wise product, represented by $\odot$, is used in the calculation of $\tilde{\boldsymbol{h}}_{t}$.

In the scenario where a time series $\boldsymbol{X}=\left\{\boldsymbol{x}_{1}, \boldsymbol{x}_{2}, \ldots, \boldsymbol{x}_{T}\right\}$ contains missing values, we use a $T$-dimensional mask vector $\boldsymbol{M}=\left\{\boldsymbol{m}_{1}, \boldsymbol{m}_{2}, \ldots, \boldsymbol{m}_{T}\right\}$ to indicate the missingness, where $\boldsymbol{m}_{t} \in \left\{0,1\right\}$ and $m_{t}$ is equal to 1 if $\boldsymbol{x}_{t}$ is revealed and 0 if $\boldsymbol{x}_{t}$ is missing. As our focus is on incomplete time series classification, there is a target label $y^{(i)}$ for the $i$-th time series $\boldsymbol{X}^{i}$ in the data set  $\mathbb{D}=\left\{\left(\boldsymbol{X}^{i}, \boldsymbol{m}^{i}, y^{i}\right)\right\}_{i=1}^{N}$.

To address the missing values in time series data, we use the GRU model as our imputation method, as shown in Fig.~\ref{Model}. At each time step $t$, the input $\boldsymbol{x}_{t}$ is encoded through GRU cells to produce the hidden state $\boldsymbol{h}_{t}$. This hidden state is then passed through a fully-connected layer to estimate the next value $\tilde{\boldsymbol{x}}_{t+1}$. The estimate is then compared to the mask vector $m_{t}$. If the input at the current time step is missing (represented by \texttt{NAN}), the previous time step's estimated value $\tilde{\boldsymbol{x}}_{t}$ is used for imputation.

We train $\tilde{\boldsymbol{x}}_{t}$ using the previous hidden state $\boldsymbol{h}_{t-1}$ to approximate the input $\boldsymbol{x}_{t}$ when it is not missing, as follows:
\begin{equation}
   \tilde{\boldsymbol{x}}_{t}=\mathbf{W}_{\mathbf{i m p}} \boldsymbol{h}_{t-1}+\boldsymbol{b},
\end{equation}
where $\mathbf{W}_{\mathbf{imp}} \in \mathbb{R}^{n \times m}$ represents a learned regression matrix, and $\boldsymbol{b}$ represents a bias term. Given that the approximation of the next value, $\boldsymbol{x}_{t}$, is trained into $\tilde{{\boldsymbol{x}}}_{t}$, it can be utilized to impute missing values. The resulting imputed and completed value is computed as follows:
\begin{equation}
   \boldsymbol{u}_{t} = m_{t} \odot \boldsymbol{x}_{t}+\left(1-m_{t}\right) \odot \tilde{\boldsymbol{x}}_{t}.
\end{equation}

The training of the GRU model uses the imputed and completed value $\boldsymbol{u}_{t}$ to finally obtain the updated formula for the GRU:
\begin{equation}
   \boldsymbol{h}_{t}=\operatorname{GRU}\left(\boldsymbol{h}_{t-1}, \boldsymbol{u}_{t} ; \mathbf{W}\right),
\end{equation}
where $\boldsymbol{h}_{t}$ denotes the hidden state vector at time step $t$, and $\mathbf{W}$ encompasses the parameters of the input-to-hidden and hidden-to-hidden transformations in Eq. (\ref{EQ1}) -  (\ref{EQ4}).

 \begin{figure}
  \centering
  \includegraphics[width=\linewidth]{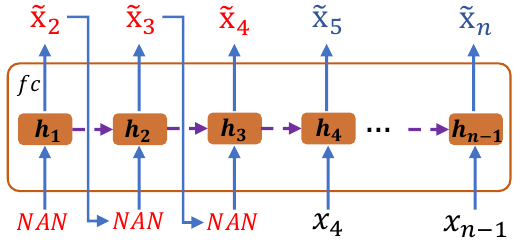}
  \caption{Illustration of temporal imputation when the first few steps contain missing values. It is evident that the estimated values $\tilde{{x}}_{2}$, $\tilde{{x}}_{3}$, and $\tilde{{x}}_{4}$ are incorrect. Furthermore, the learned hidden values ($\tilde{{h}}_{1}$ - $\tilde{{h}}_{3}$) can perpetuate the errors in the estimation of $\tilde{{x}}_{5}$.}
  \label{GRUSmodel}
  \vspace{-3mm}
\end{figure}

\subsection{Multi-Scale Feature Learning}

The information obtained from the GRU-based temporal imputation module should be viewed with a degree of skepticism, due to its limitations. As shown in Fig.~\ref{GRUSmodel}, if the initial time steps have missing values, the errors generated by the GRU-based imputation module may accumulate and result in substantial deviations in the estimated values of subsequent steps. This is due to the accumulation of hidden values ($\boldsymbol{h}_{t}$) between different time steps, which can lead to large errors in the hidden values generated from the non-missing data, causing significant deviations in the estimated values.

Therefore, the focus of our inquiry should be on identifying methods for extracting meaningful features for classification from data that is marred by the presence of noise. Furthermore, this study takes into account the possibility of MCAR (missing completely at random) missing data. In other words, the occurrence of missing values is assumed to be completely random. A particularly challenging scenario is one in which there are large amounts of missing data with a high missing rate, particularly in cases where missing values are consecutive.

To tackle the challenges mentioned earlier, this research suggests using a multi-scale representation learning (MSFL) model that combines a multi-scale mechanism and a large kernel size. This approach aims to establish correlations across different ranges while keeping computational requirements to a minimum. The multi-scale structure is designed to capture diverse features at various scales, accommodating the unpredictable nature of missing data. Additionally, the large kernel structure enables the model to capture long-range dependencies, thereby mitigating the negative impact of extensive and contiguous missing values. Moreover, to further improve the model's performance when dealing with large continuous missing data in a sizable area, we employ dilation convolution, which offers a relatively low computational burden.

As presented in Fig.~\ref{Model}, our multi-scale feature learning module consists of $N$ layers and there are $K$ kernels with different scales in each layer. The 1D CNN is utilized as the feature extractor to learn features. Given the input $\boldsymbol{X}$, the whole process of $i$-th layer can be formulated as follows: 

\begin{equation}
\begin{aligned}
& \boldsymbol{X}_k^{'} = \operatorname{Conv1d}(\boldsymbol{X}, f_k^i, d), & k \in\{1,2,\ldots,K\}, \\
& \boldsymbol{X}_k^{''} = \operatorname{ReLU}(\boldsymbol{X}_k^{'}), \\
& \boldsymbol{X}_k^{'''} = \operatorname{BN}(\boldsymbol{X}_k^{''}), \\
\end{aligned}
\end{equation}
where $f_k^i$ denotes the $k$-th kernel size in the $i$-th layer, $d$ is the dilation size, $\operatorname{BN}$ denotes the batch normalization. $\boldsymbol{X}_k^{'}$, $\boldsymbol{X}_k^{''}$, and $\boldsymbol{X}_k^{'''}$ are the output of $\operatorname{Conv1d}$ layer, $\operatorname{ReLU}$ layer, and $\operatorname{BN}$ layer for the $k$-th kernel size, respectively. 

The kernel set is defined as follows:  
\begin{equation}
    \mathbb{F}^{i}=\left\{\begin{array}{ll}\left\{\!4k + 3\right\}, & 1 \leq k \leq K, 1 \leq i \leq N-1, \\
\{1,3,5\}, &  i\!=\!N.
\end{array}\right.
\end{equation}
where $\mathbb{F}^{i}$ denotes the set of sizes of multi-scale convolutional kernels of $i$-th layer in the whole MSFL module. The $K$ hyperparameter controls the number of scales, denoted $\left\{7, 11 \ldots, 4K + 3 \right\}$, which will be analyzed sensitively in the experimental section ~\ref{experiment}.


Through the proper selection of the kernel size, we could cover many receptive fields in a range. In this way, we could implement the multi-scale feature learning from the `noisy' imputed data. As mentioned earlier, the kernel size is set larger than conventional 1D CNN to enable long-range correlations. We thus start it from 7, and then set the interval size between each kernel size to 4. This value is set to make better use of dilated convolutions. For each layer, before input enters each kernel, the  same padding is performed. The $k$ different kernels work parallel and the output of each layer is the concatenation of these $k$ different kernels.

Finally, the output features of the $N$-th layer are denoted as $\boldsymbol{X}_{N}$ and are fed into a classifier to obtain the probability distribution for each class label, which we use the softmax:
\begin{equation}
    P\left(\hat{\boldsymbol{y}}_{\boldsymbol{j}} \mid \boldsymbol{X}_{N}\right)=\frac{\exp \left(\boldsymbol{W}_{\boldsymbol{j}}^{\boldsymbol{\top}} \boldsymbol{X}_{N}\right)}{\sum_{l=1}^{K} \exp \left(\boldsymbol{W}_{\boldsymbol{l}}^{\boldsymbol{\top}} \boldsymbol{X}_{N}\right)},
    \label{EQ13}
\end{equation}
where $K$ is the number of class labels and $\left\{\boldsymbol{W}_{l}\right\}_{l=1}^{K}$ denotes the specific class weights for the softmax layer for $l$=1,\ldots,$K$. 

The classifier can be a more complex network as required by the task. We use this simple classifier because our primary goal is to demonstrate the impact of feature extraction and missing value imputation and report the results obtained. For fairness, we use this approach for all methods, our method, and the comparison methods we implement.

\subsection{Joint Learning Strategy} \label{ssec:joint_learning}

Compared with the two-stage method, our model adopts a joint learning strategy, mainly because we expect that the label information can also be utilized by the imputation model. In the joint learning process, there are two tasks: imputation and classification. Let the subscript $i$ denote the $i$-th time series sample in the dataset $\mathbb{D}$. For the imputation task, we obtain the attributed sequence vector $\tilde{\boldsymbol{X}}^{i}=\left\{\tilde{\boldsymbol{x}}_{2}^{i}, \ldots, \tilde{\boldsymbol{x}}_{T}^{i}\right\}$ for the $i$-th time series sample, which can be divided into two parts: the approximate values (observable values in sequences) and the imputed values (missing values in sequences). We compute the loss between the observed values of the input time series at the time steps without missing values and the approximate values for all time series samples. The imputation loss for all time series samples is the sum of the imputation losses for all time series samples calculated on the approximate values as follows:
\begin{equation}
    \mathcal{L}_{\mathbf{imp}}(\boldsymbol{X}, \!\tilde{\boldsymbol{X}}, \!\boldsymbol{m})=\frac{1}{Q} \sum_{i=1}^{N}\left\|\left(\boldsymbol{X}_{2: T}^{i}-\tilde{\boldsymbol{X}}^{i}\right) \odot m_{2: T}^{i}\right\|_{2}^{2},
\end{equation}
where $Q$ is the number of samples in the dataset. $\mathcal{L}_{\mathbf{imp}}$ is the mean squared error loss between the approximate and revealed values. $m_{2: T}^{i}$ masks the estimated values in the loss estimate, because there is no ground truth for the missing values.

For the classification task, we obtain the predicted probability distribution $\hat{y}^{i}$ for the $i$-th time series sample given by Eq. (\ref{EQ13}), and the loss for all time series samples can be computed as follows:
\begin{equation}
    \mathcal{L}_{\mathbf{c l s}}(y, \hat{y})=-\frac{1}{Q} \sum_{i=1}^{Q} \sum_{j=1}^{C} \mathbf{1}\left\{y^{i}=j\right\} \log \hat{y}^{i},
    \label{EQ15}
\end{equation}
where $C$ is the number of class labels. Eq. (\ref{EQ15}) is the softmax cross-entropy loss between the predicted and true labels.

Finally, we train the entire model using the loss defined as follows: 
\begin{equation}
    \mathcal{L}_{\mathbf{m o d e l}}=\alpha \mathcal{L}_{\mathbf{cls}}+\beta \mathcal{L}_{\mathbf{imp}},
\end{equation}
where $\alpha$ and $\beta$ are two hyper-parameters to adjust the weight of these above two loss functions. 

This forms an end-to-end training framework for the classification of incomplete time series. This loss function can be optimized using the back-propagation algorithm, as described in the training method in Algorithm \ref{AL1}. Our model combines the advantages of joint learning and strong feature extraction. The multi-scale feature learning module is trained to learn and classify the features of the imputed sequence and is able to provide gradients for imputation, guiding imputation in a direction that is beneficial for classification. The mask vector effectively provides supervision for each imputation. As a result, the negative impact of missing values on the classification task of our model is reduced.

\begin{algorithm}[t!]
    \SetKwInOut{Input}{\textbf{Input}}
    \SetKwInOut{Output}{\textbf{Output}} 
    \Input{Initial imputation model weight $\Phi$, Initial multi-scale feature learning module weight $\Psi$, Initial classifier weight $\Omega$, Batch\_size $B$, Training epochs $E$,  Incomplete dataset with missing values $(\boldsymbol{X}, y)$, Missing masks $\boldsymbol{m}$.} 
    \Output{Well-trained model: $\Phi$, $\Psi$, and $\Omega$.} 
    \BlankLine
     initialization\;
    \For{$e = 0;\ e < E;\ e = e + 1$}
    {
        \For{$b = 0;\ b < B;\ b = b + 1$}
        {
            \For{$t = 0;\ t < T;\ t = t + 1$}
            {
                $\boldsymbol{u}_{t}^{b}=m_{t}^{b} \odot \boldsymbol{x}_{t}^{b}+\left(1-m_{t}^{b}\right) \odot \tilde{\boldsymbol{x}}_{t}^{b}$
    
               $\tilde{\boldsymbol{x}}_{t+1}^{b} \leftarrow \Phi \left(\boldsymbol{u}_{t}^{b}\right)$
            }
    
            $\boldsymbol{U}^{b} \leftarrow\left\{\boldsymbol{u}_{1}^{b}, \cdots, \boldsymbol{u}_{t}^{b}\right\}, \tilde{\boldsymbol{x}}^{b} \leftarrow\left\{\tilde{\boldsymbol{x}}_{1}^{b}, \cdots, \tilde{\boldsymbol{x}}_{t}^{b}\right\}$
            
           $\hat{y}^{b} \leftarrow \Omega \left(\Psi \left(\boldsymbol{U}^{b}\right)\right)$
    
           $l_\mathbf{imp} \gets  \mathcal{L}_{\mathbf{imp}}\left(X^{b}, \tilde{\boldsymbol{x}}^b, M^b \right)$ 
    
           $l_{\mathbf{c l s}} \gets \mathcal{L}_{\mathbf{c l s}}\left(y^{b}, \hat{y}^{b}\right)$ 
    
           $\mathcal{L}_{\mathbf{total}}= \alpha l_{\mathbf{c l s}} +\beta l_{\mathbf{c l s}}$
    
           $\Phi \gets \Phi - \mathcal{L}_{\mathbf{total}}\left(\Phi\right)$
    
           $\Psi \gets \Psi - \mathcal{L}_{\mathbf{total}}\left(\Psi\right)$
    
           $\Omega \gets \Omega - \mathcal{L}_{\mathbf{total}}\left(\Omega\right)$
        }
    }
    \caption{Training algorithm\label{AL1}} 
\end{algorithm}
\vspace{-9pt}

\section{Experiments} \label{experiment}
The model is implemented using the deep learning framework, Pytorch (v.1.10.0). All experiments are performed on a server equipped with an NVIDIA Tesla V100 GPU. We set the learning rate to $3\times 10^{-4}$ for Adam optimizer. The batch size is set to $64$, but for the datasets with smaller sample sizes, we use $16$, such as $CBF$ $ArrawHead$ etc. The training epoch is set to $100$. 

\subsection{Datasets}

The following three different kinds of datasets are used in our experiments:
\begin{itemize}[leftmargin=*]
\item \textbf{Univariate time series}: We select 68 public univariate time series data from the UCR Time Series Archive \cite{dau2019ucr}. The 68 datasets are from various domains and have different numbers of training/testing samples. Following the previous work \cite{ma_ajrnn}, there are four missing rates varied from $20\%$ to $80\%$, with increments of $20\%$ per dataset; Also, classification accuracy is used as the metric. 

\item \textbf{Multivariate time series}: The physical activity monitoring (PAM) dataset \cite{PAM2Dataset} is utilized for the model performance evaluation. The PAM dataset uses three inertial measurement units to measure the daily activities of nine participants. It is modified to fit an incomplete time series classification scenario. Due to the short length of sensor readings, the ninth participant is excluded. Continuous signals are divided into samples with a 600 time window and 50\% overlap. The original PAM dataset has 18 daily activities, but those associated with less than 500 samples are excluded, leaving eight activities. The modified PAM dataset contains 5,333 segments of sensory samples. Each sample is measured by 17 sensors, contains 600 consecutive observations, and has a sampling frequency of 100 Hz. To make the time series missing, 60\% of the observations are randomly removed while maintaining fairness by keeping the same observations removed for all experimental setups and methods. PAM is labeled by eight classes, each representing a different daily activity. PAM does not include static attributes, and samples are roughly balanced across all eight classes. Following the previous work \cite{zhang2021graph}, accuracy, precision, recall, and F1 are used as the metrics. 

\item \textbf{Real-world dataset}: In order to evaluate performance, four real-world datasets with missing information, namely DodgerLpDay, DodgerLpGame, DodgerLpWend, and MelPedestrian, are obtained from the UCR archive extension. These datasets contain missing values that occurred naturally during the collection process, as opposed to being artificially introduced. Thus, these datasets provide an accurate representation of the challenges and complexities associated with working with incomplete data in real-world scenarios. For the DodgerLpDay dataset, approximately 14\% of the information is missing in the training dataset, and 4\% is missing in the testing dataset. Similarly, for the DodgerLpGame dataset, approximately 15\% of the information is missing in the training dataset, and 8\% is missing in the testing dataset. Additionally, for the DodgerLpWend dataset, approximately 10\% of the information is missing in the training dataset, and 9\% is missing in the testing dataset. Lastly, for the MelPedestrian dataset, approximately 5\% of the information is missing in both the training and testing datasets. The classification accuracy is used as the evaluation metric according to the previous work \cite{ma_ajrnn}.

\end{itemize}

\subsection{Results on Univariate Dataset}
\begin{table*}
\renewcommand\arraystretch{1.2}
\tabcolsep=3mm
\centering
\caption{Comparison results of ITSC in 68 univariate datasets.}
\resizebox{\linewidth}{!}{ 
{ 
\fontsize{50}{54}\selectfont 
\begin{tabular}{c|ccccc|ccccc|ccccc|ccccc} 
\hline
\diagbox{Dataset}{Acc}{Ratio}            & \multicolumn{5}{c|}{20\%}                                                                                                                                       & \multicolumn{5}{c|}{40\%}                                                                                                                                        & \multicolumn{5}{c|}{60\%}                                                                                                                                       & \multicolumn{5}{c}{80\%}                                                                                                                                        \\ 
\hline
                             & BRITS                     & GRUD                       & Raindrop                    & AJRNN                      & Ours                                        & BRITS                     & GRUD                       & Raindrop                    & AJRNN                      & Ours                                         & BRITS                    & GRUD                       & Raindrop                   & AJRNN                     & Ours                                         & BRITS                     & GRUD                       & Raindrop                    & AJRNN                      & Ours                                         \\ 
\cline{2-21}
50words                            & 0.367          & 0.125 & 0.490          & 0.736          & \textbf{0.807} & 0.343 & 0.539 & 0.407          & 0.711          & \textbf{0.820} & 0.292 & 0.132 & 0.314          & 0.679          & \textbf{0.804} & 0.268 & 0.106 & 0.185    & 0.580          & \textbf{0.758} \\
Adiac                              & 0.217          & 0.023 & 0.143          & 0.571          & \textbf{0.798} & 0.207 & 0.041 & 0.043          & 0.539          & \textbf{0.788} & 0.130 & 0.021 & 0.018          & 0.416          & \textbf{0.742} & 0.133 & 0.028 & 0.054    & 0.301          & \textbf{0.453} \\
ArrowHead                          & 0.303          & 0.406 & 0.606          & 0.699          & \textbf{0.897} & 0.406 & 0.326 & 0.366          & 0.667          & \textbf{0.886} & 0.417 & 0.314 & 0.549          & 0.606          & \textbf{0.851} & 0.383 & 0.337 & 0.463    & 0.530          & \textbf{0.714} \\
CBF                                & 0.598          & 0.373 & 0.827          & 0.993          & \textbf{0.998} & 0.582 & 0.377 & 0.766          & 0.979          & \textbf{0.993} & 0.441 & 0.376 & 0.511          & 0.934          & \textbf{0.978} & 0.398 & 0.354 & 0.331    & 0.814          & \textbf{0.916} \\
Chlorine                           & 0.542          & 0.533 & 0.559          & 0.565          & \textbf{0.566} & 0.535 & 0.533 & 0.533          & 0.551          & \textbf{0.554} & 0.533 & 0.533 & 0.482          & 0.549          & \textbf{0.554} & 0.533 & 0.532 & 0.362    & 0.542          & \textbf{0.550} \\
CinC\_ECG                          & 0.298          & 0.273 & \textbf{0.891} & 0.450          & 0.700          & 0.269 & 0.241 & \textbf{0.828} & 0.425          & 0.748          & 0.320 & 0.249 & \textbf{0.684} & 0.395          & 0.673          & 0.248 & 0.247 & 0.425    & 0.371          & \textbf{0.680} \\
Computers                          & 0.636          & 0.560 & 0.560          & 0.672          & \textbf{0.696} & 0.652 & 0.516 & 0.572          & 0.667          & \textbf{0.688} & 0.620 & 0.500 & 0.500          & 0.659          & \textbf{0.696} & 0.644 & 0.544 & 0.500    & 0.645          & \textbf{0.660} \\
Cricket\_X                         & 0.144          & 0.123 & 0.385          & 0.617          & \textbf{0.836} & 0.236 & 0.105 & 0.280          & 0.604          & \textbf{0.839} & 0.231 & 0.087 & 0.239          & 0.528          & \textbf{0.823} & 0.221 & 0.082 & 0.108    & 0.445          & \textbf{0.718} \\
Cricket\_Y                         & 0.221          & 0.089 & 0.349          & 0.642          & \textbf{0.831} & 0.203 & 0.141 & 0.405          & 0.617          & \textbf{0.815} & 0.285 & 0.062 & 0.203          & 0.566          & \textbf{0.782} & 0.167 & 0.115 & 0.146    & 0.496          & \textbf{0.687} \\
Cricket\_Z                         & 0.236          & 0.133 & 0.439          & 0.613          & \textbf{0.859} & 0.131 & 0.077 & 0.146          & 0.598          & \textbf{0.849} & 0.182 & 0.087 & 0.154          & 0.545          & \textbf{0.818} & 0.218 & 0.095 & 0.136    & 0.343          & \textbf{0.710} \\
DiaSizReduc                        & 0.301          & 0.297 & 0.657          & 0.819          & \textbf{0.925} & 0.301 & 0.177 & 0.438          & 0.800          & \textbf{0.886} & 0.294 & 0.108 & 0.363          & 0.762          & \textbf{0.801} & 0.301 & 0.288 & 0.301    & 0.470          & \textbf{0.663} \\
DistlPhxAgeGp                      & 0.803          & 0.785 & 0.813          & \textbf{0.849} & 0.830          & 0.793 & 0.793 & 0.765          & \textbf{0.840} & 0.828          & 0.755 & 0.665 & 0.715          & \textbf{0.810} & 0.805          & 0.663 & 0.643 & 0.670    & 0.712          & \textbf{0.758} \\
DistalPhxCorr                      & 0.680          & 0.470 & 0.500          & 0.778          & \textbf{0.782} & 0.558 & 0.600 & 0.557          & 0.720          & \textbf{0.787} & 0.653 & 0.612 & 0.585          & 0.677          & \textbf{0.763} & 0.655 & 0.370 & 0.540    & 0.630          & \textbf{0.722} \\
DistalPhxTW                        & 0.790          & 0.628 & 0.770          & \textbf{0.799} & 0.788          & 0.770 & 0.710 & 0.718          & \textbf{0.780} & 0.768          & 0.688 & 0.618 & 0.645          & 0.754          & \textbf{0.758} & 0.623 & 0.520 & 0.518    & 0.708          & \textbf{0.725} \\
Earthquakes                        & 0.823          & 0.823 & 0.562          & \textbf{0.836} & 0.823          & 0.820 & 0.823 & 0.745          & \textbf{0.832} & 0.823          & 0.817 & 0.817 & 0.702          & \textbf{0.830} & 0.823          & 0.820 & 0.823 & 0.748    & \textbf{0.826} & 0.823          \\
ECG5000                            & 0.932          & 0.939 & 0.922          & 0.937          & \textbf{0.946} & 0.931 & 0.802 & 0.929          & 0.935          & \textbf{0.946} & 0.911 & 0.811 & 0.918          & 0.933          & \textbf{0.944} & 0.919 & 0.649 & 0.884    & 0.928          & \textbf{0.939} \\
ECGFiveDays                        & 0.497          & 0.664 & 0.717          & 0.976          & \textbf{0.994} & 0.497 & 0.623 & 0.497          & 0.866          & \textbf{0.987} & 0.497 & 0.507 & 0.497          & 0.802          & \textbf{0.944} & 0.497 & 0.497 & 0.497    & 0.697          & \textbf{0.762} \\
ElecDev                            & 0.684          & 0.644 & 0.601          & \textbf{0.724} & 0.691          & 0.607 & 0.664 & 0.571          & \textbf{0.691} & 0.663          & 0.622 & 0.592 & 0.564          & \textbf{0.661} & 0.607          & 0.530 & 0.577 & 0.500    & \textbf{0.580} & 0.557          \\
FaceAll                            & \textbf{0.801} & 0.511 & 0.797          & 0.764          & 0.784          & 0.726 & 0.583 & 0.706          & 0.749          & \textbf{0.780} & 0.653 & 0.383 & 0.472          & 0.651          & \textbf{0.745} & 0.451 & 0.248 & 0.299    & 0.446          & \textbf{0.637} \\
FacesUCR                           & 0.405          & 0.242 & 0.642          & 0.818          & \textbf{0.868} & 0.454 & 0.215 & 0.412          & 0.739          & \textbf{0.807} & 0.349 & 0.270 & 0.329          & 0.546          & \textbf{0.689} & 0.239 & 0.145 & 0.185    & 0.337          & \textbf{0.453} \\
FISH                               & 0.217          & 0.149 & 0.480          & 0.653          & \textbf{0.971} & 0.166 & 0.120 & 0.417          & 0.541          & \textbf{0.977} & 0.166 & 0.137 & 0.183          & 0.345          & \textbf{0.977} & 0.183 & 0.120 & 0.183    & 0.265          & \textbf{0.920} \\
FordA                              & 0.674          & 0.513 & 0.710          & 0.924          & \textbf{0.954} & 0.696 & 0.509 & 0.642          & 0.918          & \textbf{0.951} & 0.492 & 0.487 & 0.623          & 0.906          & \textbf{0.945} & 0.527 & 0.512 & 0.566    & 0.628          & \textbf{0.908} \\
FordB                              & 0.570          & 0.505 & 0.522          & 0.900          & \textbf{0.927} & 0.566 & 0.501 & 0.570          & 0.853          & \textbf{0.927} & 0.584 & 0.506 & 0.493          & 0.726          & \textbf{0.924} & 0.520 & 0.510 & 0.481    & 0.543          & \textbf{0.888} \\
Ham                                & 0.581          & 0.514 & 0.724          & 0.702          & \textbf{0.857} & 0.581 & 0.514 & 0.800          & 0.686          & \textbf{0.791} & 0.667 & 0.524 & 0.667          & 0.670          & \textbf{0.800} & 0.543 & 0.457 & 0.552    & 0.654          & \textbf{0.752} \\
HandOutlines                       & 0.645          & 0.462 & 0.855          & 0.684          & \textbf{0.871} & 0.638 & 0.638 & 0.772          & 0.667          & \textbf{0.840} & 0.638 & 0.631 & 0.841          & 0.646          & \textbf{0.860} & 0.638 & 0.566 & 0.763    & 0.642          & \textbf{0.868} \\
Haptics                            & 0.331          & 0.308 & 0.390          & 0.372          & \textbf{0.533} & 0.347 & 0.175 & 0.357          & 0.358          & \textbf{0.523} & 0.338 & 0.192 & 0.328          & 0.351          & \textbf{0.539} & 0.279 & 0.221 & 0.308    & 0.329          & \textbf{0.546} \\
InlineSkate                        & 0.180          & 0.195 & 0.211          & 0.273          & \textbf{0.422} & 0.193 & 0.164 & 0.202          & 0.246          & \textbf{0.387} & 0.196 & 0.138 & 0.184          & 0.239          & \textbf{0.347} & 0.167 & 0.182 & 0.175    & 0.224          & \textbf{0.331} \\
InsWngSnd                          & 0.531          & 0.097 & 0.579          & 0.559          & \textbf{0.642} & 0.515 & 0.108 & 0.536          & 0.547          & \textbf{0.637} & 0.520 & 0.099 & 0.435          & 0.532          & \textbf{0.618} & 0.454 & 0.093 & 0.293    & 0.486          & \textbf{0.569} \\
ItalyPowDem                        & 0.941          & 0.926 & 0.902          & \textbf{0.951} & 0.946          & 0.906 & 0.655 & 0.825          & \textbf{0.931} & 0.913          & 0.830 & 0.556 & 0.752          & 0.853          & \textbf{0.878} & 0.712 & 0.632 & 0.499    & 0.732          & \textbf{0.786} \\
LrgKitApp                          & 0.451          & 0.443 & 0.448          & 0.879          & \textbf{0.891} & 0.405 & 0.368 & 0.419          & 0.855          & \textbf{0.880} & 0.437 & 0.355 & 0.408          & 0.794          & \textbf{0.864} & 0.421 & 0.333 & 0.416    & 0.660          & \textbf{0.757} \\
MALLAT                             & 0.123          & 0.126 & 0.563          & 0.550          & \textbf{0.936} & 0.123 & 0.123 & 0.615          & 0.505          & \textbf{0.944} & 0.123 & 0.122 & 0.287          & 0.439          & \textbf{0.899} & 0.119 & 0.128 & 0.278    & 0.198          & \textbf{0.881} \\
MediImgs                           & 0.628          & 0.562 & 0.578          & 0.694          & \textbf{0.736} & 0.600 & 0.508 & 0.488          & 0.663          & \textbf{0.713} & 0.554 & 0.504 & 0.453          & 0.601          & \textbf{0.682} & 0.517 & 0.493 & 0.482    & 0.557          & \textbf{0.636} \\
MidPhxAgeGp                        & 0.398          & 0.270 & 0.270          & 0.785          & \textbf{0.798} & 0.748 & 0.270 & 0.270          & 0.764          & \textbf{0.775} & 0.270 & 0.270 & 0.270          & 0.753          & \textbf{0.765} & 0.270 & 0.258 & 0.270    & 0.367          & \textbf{0.653} \\
MidPhxCorr                         & 0.478          & 0.632 & 0.460          & 0.729          & \textbf{0.740} & 0.463 & 0.647 & 0.492          & 0.649          & \textbf{0.717} & 0.483 & 0.630 & 0.468          & \textbf{0.647} & \textbf{0.647} & 0.645 & 0.622 & 0.533    & 0.621          & \textbf{0.647} \\
MidPhxTW                           & 0.622          & 0.597 & 0.609          & 0.637          & \textbf{0.647} & 0.612 & 0.531 & 0.499          & \textbf{0.632} & 0.629          & 0.531 & 0.406 & 0.459          & 0.612          & \textbf{0.632} & 0.386 & 0.293 & 0.378    & 0.476          & \textbf{0.589} \\
MoteStrain                         & 0.800          & 0.744 & 0.784          & 0.818          & \textbf{0.841} & 0.809 & 0.733 & 0.812          & 0.811          & \textbf{0.813} & 0.754 & 0.597 & 0.743          & 0.804          & \textbf{0.805} & 0.739 & 0.542 & 0.461    & 0.734          & \textbf{0.765} \\
NonInv\_Thor1                       & 0.578          & 0.040 & 0.623          & 0.847          & \textbf{0.926} & 0.308 & 0.029 & 0.416          & 0.811          & \textbf{0.924} & 0.534 & 0.030 & 0.021          & 0.762          & \textbf{0.906} & 0.144 & 0.024 & 0.109    & 0.647          & \textbf{0.850} \\
NonInv\_Thor2                       & 0.738          & 0.156 & 0.645          & 0.891          & \textbf{0.942} & 0.721 & 0.020 & 0.417          & 0.855          & \textbf{0.937} & 0.626 & 0.021 & 0.354          & 0.801          & \textbf{0.924} & 0.542 & 0.022 & 0.163    & 0.728          & \textbf{0.887} \\
OSULeaf                            & 0.417          & 0.182 & 0.426          & 0.675          & \textbf{0.864} & 0.360 & 0.211 & 0.455          & 0.612          & \textbf{0.880} & 0.405 & 0.211 & 0.388          & 0.600          & \textbf{0.880} & 0.306 & 0.182 & 0.273    & 0.515          & \textbf{0.781} \\
PhalOutCorr                        & 0.711          & 0.613 & 0.273          & 0.775          & \textbf{0.824} & 0.620 & 0.613 & 0.617          & 0.730          & \textbf{0.800} & 0.619 & 0.613 & 0.608          & 0.695          & \textbf{0.780} & 0.618 & 0.613 & 0.614    & 0.629          & \textbf{0.699} \\
Phoneme                            & 0.134          & 0.115 & 0.087          & 0.148          & \textbf{0.224} & 0.138 & 0.124 & 0.098          & 0.144          & \textbf{0.242} & 0.116 & 0.122 & 0.077          & 0.137          & \textbf{0.228} & 0.123 & 0.012 & 0.076    & 0.135          & \textbf{0.196} \\
Plane                              & 0.524          & 0.276 & 0.962          & 0.997          & \textbf{1.000} & 0.533 & 0.114 & 0.886          & 0.984          & \textbf{0.991} & 0.362 & 0.105 & 0.857          & 0.959          & \textbf{0.971} & 0.105 & 0.095 & 0.419    & 0.838          & \textbf{0.905} \\
ProxPhxAgeGp                       & 0.849          & 0.805 & 0.829          & \textbf{0.876} & 0.868          & 0.815 & 0.805 & 0.790          & 0.870          & \textbf{0.873} & 0.849 & 0.834 & 0.678          & 0.865          & \textbf{0.873} & 0.795 & 0.649 & 0.581    & 0.850          & \textbf{0.854} \\
ProxPhxCorr                        & 0.687          & 0.722 & 0.739          & 0.825          & \textbf{0.845} & 0.691 & 0.684 & 0.708          & 0.755          & \textbf{0.814} & 0.687 & 0.680 & 0.643          & 0.738          & \textbf{0.790} & 0.684 & 0.684 & 0.619    & 0.725          & \textbf{0.760} \\
ProxPhxTW                          & 0.708          & 0.450 & 0.750          & 0.808          & \textbf{0.820} & 0.710 & 0.450 & 0.658          & 0.789          & \textbf{0.798} & 0.703 & 0.450 & 0.590          & 0.771          & \textbf{0.780} & 0.507 & 0.298 & 0.460    & 0.731          & \textbf{0.750} \\
RefrgDev                           & 0.528          & 0.339 & 0.365          & 0.542          & \textbf{0.552} & 0.416 & 0.301 & 0.296          & \textbf{0.528} & \textbf{0.528} & 0.507 & 0.371 & 0.339          & 0.515          & \textbf{0.525} & 0.456 & 0.344 & 0.361    & 0.470          & \textbf{0.525} \\
ScreenType                         & 0.400          & 0.360 & 0.405          & \textbf{0.469} & 0.464          & 0.408 & 0.307 & 0.413          & 0.459          & \textbf{0.475} & 0.403 & 0.333 & 0.411          & 0.453          & \textbf{0.456} & 0.405 & 0.368 & 0.389    & 0.444          & \textbf{0.448} \\
ShapesAll                          & 0.402          & 0.045 & 0.560          & 0.763          & \textbf{0.857} & 0.317 & 0.277 & 0.363          & 0.734          & \textbf{0.862} & 0.165 & 0.155 & 0.230          & 0.655          & \textbf{0.845} & 0.338 & 0.077 & 0.058    & 0.574          & \textbf{0.810} \\
SmlKitApp                          & 0.560          & 0.333 & 0.480          & 0.705          & \textbf{0.805} & 0.568 & 0.325 & 0.392          & 0.678          & \textbf{0.792} & 0.547 & 0.333 & 0.397          & 0.652          & \textbf{0.744} & 0.381 & 0.341 & 0.413    & 0.570          & \textbf{0.667} \\
SonyRobot\_Sur                      & 0.429          & 0.429 & 0.429          & \textbf{0.836} & 0.717          & 0.429 & 0.429 & 0.429          & \textbf{0.796} & 0.659          & 0.429 & 0.429 & 0.429          & \textbf{0.748} & 0.669          & 0.429 & 0.429 & 0.429    & \textbf{0.666} & 0.599          \\
SonyRobot\_Sur2                     & 0.662          & 0.652 & \textbf{0.830} & 0.818          & 0.789          & 0.745 & 0.659 & \textbf{0.796} & 0.791          & 0.758          & 0.626 & 0.633 & 0.765          & \textbf{0.769} & 0.738          & 0.617 & 0.620 & 0.617    & 0.713          & \textbf{0.720} \\
StarLitCurs                        & 0.835          & 0.823 & 0.870          & 0.972          & \textbf{0.979} & 0.839 & 0.838 & 0.825          & 0.967          & \textbf{0.978} & 0.835 & 0.810 & 0.844          & 0.954          & \textbf{0.979} & 0.831 & 0.640 & 0.824    & 0.902          & \textbf{0.977} \\
Strawberry                         & 0.682          & 0.643 & 0.832          & 0.917          & \textbf{0.949} & 0.736 & 0.643 & 0.710          & 0.901          & \textbf{0.930} & 0.649 & 0.636 & 0.677          & 0.854          & \textbf{0.912} & 0.641 & 0.643 & 0.591    & 0.729          & \textbf{0.842} \\
SwedishLeaf                        & 0.722          & 0.592 & 0.530          & 0.895          & \textbf{0.922} & 0.662 & 0.669 & 0.451          & 0.867          & \textbf{0.923} & 0.538 & 0.066 & 0.301          & 0.817          & \textbf{0.877} & 0.405 & 0.326 & 0.171    & 0.679          & \textbf{0.747} \\
Symbols                            & 0.466          & 0.245 & 0.574          & 0.867          & \textbf{0.973} & 0.360 & 0.298 & 0.538          & 0.838          & \textbf{0.974} & 0.280 & 0.166 & 0.619          & 0.788          & \textbf{0.960} & 0.174 & 0.230 & 0.180    & 0.726          & \textbf{0.861} \\
Syn\_Contr                          & 0.967          & 0.907 & 0.880          & \textbf{0.987} & \textbf{0.987} & 0.877 & 0.920 & 0.727          & \textbf{0.973} & 0.970          & 0.773 & 0.763 & 0.597          & 0.913          & \textbf{0.943} & 0.590 & 0.660 & 0.427    & 0.743          & \textbf{0.787} \\
ToeSegmtion1                       & 0.474          & 0.526 & 0.570          & 0.847          & \textbf{0.939} & 0.513 & 0.465 & 0.518          & 0.833          & \textbf{0.921} & 0.509 & 0.548 & 0.518          & 0.756          & \textbf{0.908} & 0.412 & 0.509 & 0.522    & 0.582          & \textbf{0.847} \\
Two\_Patterns                      & 0.590          & 0.366 & 0.964          & \textbf{1.000} & \textbf{1.000} & 0.988 & 0.302 & 0.874          & \textbf{1.000} & \textbf{1.000} & 0.532 & 0.262 & 0.700          & 0.992          & \textbf{0.996} & 0.489 & 0.553 & 0.427    & \textbf{0.909} & 0.898          \\
TwoLeadECG                         & 0.625          & 0.538 & 0.656          & 0.898          & \textbf{0.921} & 0.591 & 0.522 & 0.587          & \textbf{0.879} & 0.831          & 0.517 & 0.492 & 0.601          & \textbf{0.751} & 0.719          & 0.543 & 0.500 & 0.583    & 0.607          & \textbf{0.654} \\
UWavGest\_X                         & 0.475          & 0.653 & 0.708          & 0.800          & \textbf{0.815} & 0.398 & 0.208 & 0.711          & 0.794          & \textbf{0.821} & 0.529 & 0.195 & 0.660          & 0.786          & \textbf{0.812} & 0.356 & 0.692 & 0.521    & 0.777          & \textbf{0.812} \\
UWavGest\_Y                         & 0.420          & 0.559 & 0.646          & 0.721          & \textbf{0.743} & 0.470 & 0.424 & 0.599          & 0.709          & \textbf{0.752} & 0.503 & 0.384 & 0.566          & 0.697          & \textbf{0.739} & 0.494 & 0.306 & 0.513    & 0.668          & \textbf{0.716} \\
UWavGest\_Z                         & 0.526          & 0.296 & 0.624          & 0.732          & \textbf{0.768} & 0.460 & 0.311 & 0.602          & 0.727          & \textbf{0.753} & 0.336 & 0.574 & 0.586          & 0.720          & \textbf{0.761} & 0.414 & 0.609 & 0.513    & 0.707          & \textbf{0.759} \\
UWavGestAll                        & 0.383          & 0.179 & 0.930          & \textbf{0.934} & 0.922          & 0.286 & 0.315 & 0.876          & \textbf{0.926} & 0.918          & 0.388 & 0.235 & 0.883          & \textbf{0.918} & 0.913          & 0.348 & 0.248 & 0.790    & 0.902          & \textbf{0.918} \\
Wafer                              & 0.989          & 0.892 & 0.992          & \textbf{0.996} & \textbf{0.996} & 0.954 & 0.892 & 0.988          & 0.994          & \textbf{0.996} & 0.961 & 0.892 & 0.983          & 0.991          & \textbf{0.994} & 0.948 & 0.892 & 0.968    & 0.980          & \textbf{0.984} \\
WordSynms                          & 0.332          & 0.447 & 0.426          & 0.594          & \textbf{0.647} & 0.260 & 0.187 & 0.329          & 0.578          & \textbf{0.677} & 0.345 & 0.219 & 0.290          & 0.551          & \textbf{0.655} & 0.249 & 0.182 & 0.241    & 0.496          & \textbf{0.547} \\
Worms                              & 0.403          & 0.381 & 0.293          & 0.483          & \textbf{0.641} & 0.403 & 0.370 & 0.260          & 0.455          & \textbf{0.630} & 0.420 & 0.354 & 0.381          & 0.438          & \textbf{0.580} & 0.425 & 0.387 & 0.370    & 0.425          & \textbf{0.591} \\
WormsToCla                         & 0.591          & 0.580 & 0.619          & 0.641          & \textbf{0.740} & 0.569 & 0.536 & 0.553          & 0.632          & \textbf{0.751} & 0.569 & 0.470 & 0.564          & 0.626          & \textbf{0.751} & 0.580 & 0.558 & 0.519    & 0.602          & \textbf{0.751} \\
Yoga                               & 0.626          & 0.528 & 0.648          & 0.756          & \textbf{0.879} & 0.628 & 0.532 & 0.607          & 0.722          & \textbf{0.858} & 0.591 & 0.526 & 0.607          & 0.698          & \textbf{0.841} & 0.572 & 0.531 & 0.553    & 0.602          & \textbf{0.790} \\
\hline
AVG Acc                            & 0.542          & 0.446 & 0.610          & 0.750          & \textbf{0.815} & 0.532 & 0.430 & 0.557          & 0.725          & \textbf{0.804} & 0.502 & 0.394 & 0.503          & 0.685          & \textbf{0.783} & 0.455 & 0.388 & 0.423    & 0.600          & \textbf{0.725}   \\
\hline
\end{tabular}
}
}
\label{univeriate_baseline}
\vspace{-5pt}
\end{table*}

Four state-of-the-art methods: In Section ~\ref{related}, GRU-D \cite{che2018recurrent}, BRITS \cite{cao2018brits}, Raindrop \cite{zhang2021graph}, and AJRNN \cite{ma_ajrnn}, are used for comparisons. All models are implemented based on their codebases and all of the training settings follow the original papers. All datasets used in this section are processed according to \cite{ma_ajrnn}.

As demonstrated in Table~\ref{univeriate_baseline}, our model achieved superior performance across all datasets and missing ratios in terms of average accuracy compared to other models. The closest competitor was AJRNN. 
Out of the remaining three models---Raindrop, GRUD, and BRITS---they exhibited comparable performance, with Raindrop slightly outperforming the other two. Note that the performance of all methods tended to decrease as the missing ratio increased, potentially due to the increased difficulty of imputing missing values. Specifically, at 20\% missing ratio, our proposed method outperformed BRITS by 50.37\%, GRUD by 82.74\%, Raindrop by 33.61\% and AJRNN by 8.67\%. Similarly, at 40\% missing ratio, our proposed method outperformed BRITS by 51.13\%, GRUD by 86.98\%, Raindrop by 44.34\% and AJRNN by 10.90\%. At 60\% missing ratio, our proposed method outperformed BRITS by 55.98\%, GRUD by 98.73\%, Raindrop by 55.67\% and AJRNN by 14.31\%. Finally, at 80\% missing ratio, our proposed method outperformed BRITS by 59.34\%, GRUD by 86.86\%, Raindrop by 71.39\% and AJRNN by 20.83\%.

These methods used for comparison may be limited in their ability to accurately classify due to only utilizing the most salient features in their models. For example, the state-of-the-art model AJRNN suffers from errors in its network-estimated missing values. As these estimated values are fed into the network, the errors accumulate, thereby compromising the model's classification capability. This is particularly problematic as the classification model incorporated into the framework relies on the final hidden state of RNN. Although these errors do not accumulate rapidly after they are input into the discriminator, the RNN framework is still affected by these errors, resulting in a negative impact on the classification results. In this context, we posit that the RNN structure for predicting the imputation of missing time series data, based on the features of observable values and the complete sequence after imputation, holds greater importance for downstream classification tasks, regardless of whether the input consists of observable or imputed values. However, the proposed method addresses this limitation by incorporating a specifically designed feature learning module to compensate for errors caused by imputation. This module improves classification performance by extracting a more comprehensive set of features, effectively reducing the negative impact of missing data on the classification task. Furthermore, we also observe that the training difficulty is reduced compared to AJRNN which uses the adversarial training fashion. 

\begin{figure}
     \centering
     \begin{subfigure}[b]{0.225\textwidth}
         \centering
         \includegraphics[width=\textwidth]{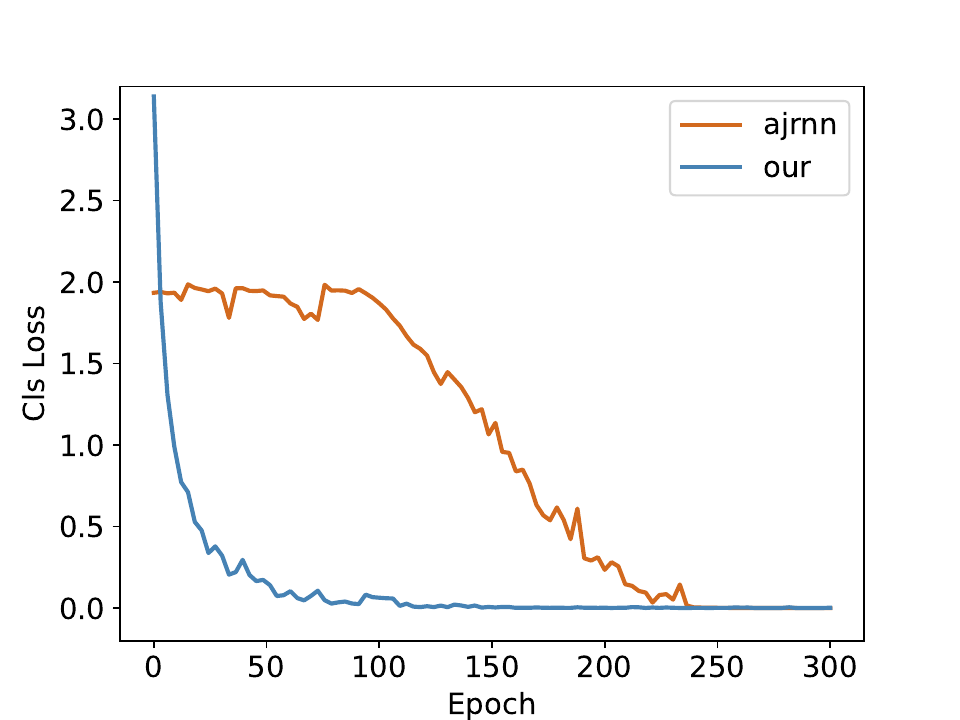}
         \caption{20\%}
     \end{subfigure}
     \hfill
     \begin{subfigure}[b]{0.225\textwidth}
         \centering
    \includegraphics[width=\textwidth]{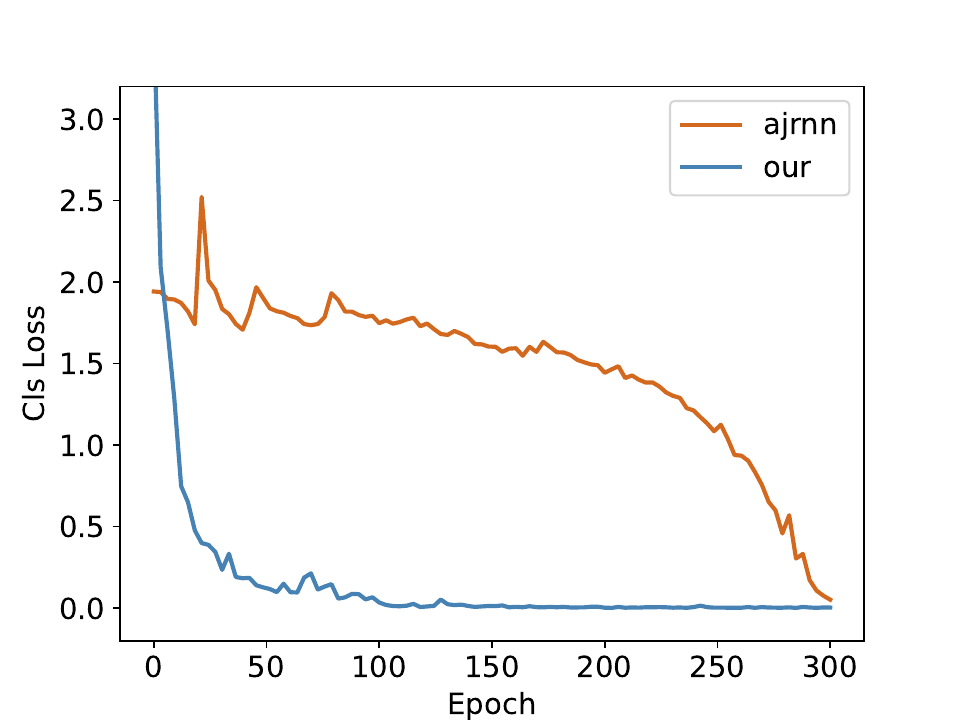}
         \caption{40\%}
     \end{subfigure}

     \begin{subfigure}[b]{0.225\textwidth}
         \centering
         \includegraphics[width=\textwidth]{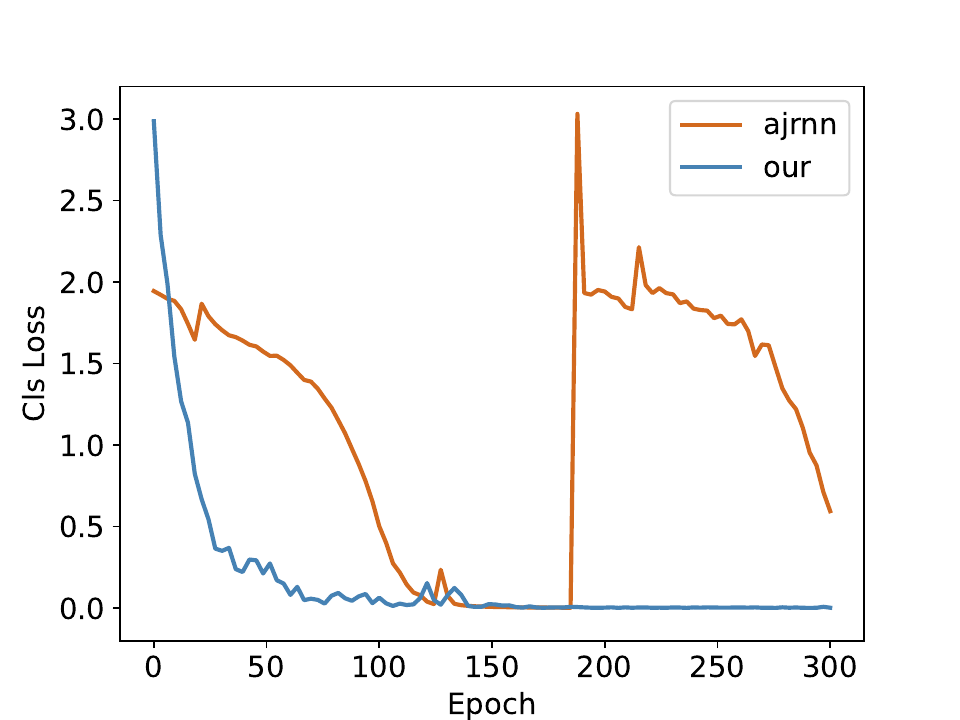}
         \caption{60\%}
     \end{subfigure}
     \hfill
     \begin{subfigure}[b]{0.225\textwidth}
         \centering
         \includegraphics[width=\textwidth]{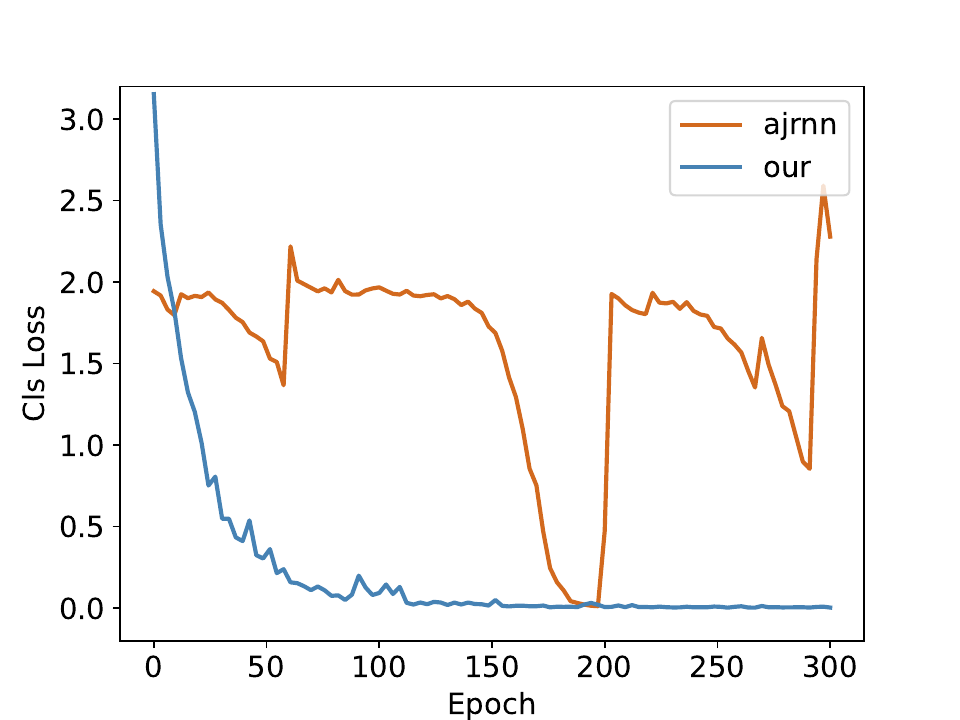}
         \caption{80\%}
     \end{subfigure}     
        \caption{The classification loss convergence during training.}
        \label{fig:LossComp}
\end{figure}

In addition, we compare the performance of our model and AJRNN model during the training in the FISH dataset, from the convergence trend of classification loss, training time consumption level. From the convergence trend in Fig.~\ref{fig:LossComp}, we can see that the AJRNN training process is still affected by noise and discriminators, which affects the classification, and our model is a stable and flat convergence process. And the same experimental environment, using the best parameter configuration of the model, we take 1.95s per epoch and AJRNN takes 3.54s per epoch. The simplicity of our model at the training level is clearly due to AJRNN.

\subsection{Results on Multivariate Dataset}
\begin{table}[!htb]
\renewcommand\arraystretch{1.2}
\scriptsize
\centering
\caption{Classification performance on samples with a fixed set of left-out sensors or random missing sensors on the PAM dataset (A: accuracy; P: precision; R: recall).}
\label{exp_PAM}
\resizebox{\linewidth}{!}{ 
{\fontsize{30}{32}\selectfont
\begin{tabular}{c|llllllllll} 
\hline
\multirow{2}{*}{Ratio} & \multicolumn{1}{c}{\multirow{2}{*}{Methods}} & \multicolumn{4}{c}{Fixed sensor out}                                                              &  & \multicolumn{4}{c}{Random sensor out}                                                              \\ 
\cline{3-6}\cline{8-11}
                       & \multicolumn{1}{c}{}                         & A                      & P                      & R                      & F1                     &  & A                      & P                      & R                      & F1                      \\ 
\cline{1-6}\cline{8-11}
\multirow{7}{*}{10\%}  & Transformer                                  & 0.603                  & 0.578                  & 0.598                  & 0.572                  &  & 0.609                  & 0.584                  & 0.591                  & 0.569                   \\
                       & Trans-mean                                   & 0.604                  & 0.618                  & 0.602                  & 0.580                  &  & 0.624                  & 0.596                  & 0.637                  & 0.627                   \\
                       & GRU-D                                        & 0.654                  & 0.726                  & 0.643                  & 0.636                  &  & 0.684                  & 0.742                  & 0.708                  & 0.720                   \\
                       & SeFT                                         & 0.589                  & 0.625                  & 0.596                  & 0.596                  &  & 0.400                  & 0.408                  & 0.410                  & 0.399                   \\
                       & mTAND                                        & 0.588                  & 0.595                  & 0.644                  & 0.618                  &  & 0.534                  & 0.548                  & 0.570                  & 0.559                   \\
                       & RAINDROP                                     & 0.772                  & 0.823                  & 0.784                  & 0.752                  &  & 0.767                  & 0.799                  & 0.779                  & 0.786                   \\
                       & Ours                                         & \textbf{0.900} & 
                       \textbf{0.923} & 
                       \textbf{0.906} & 
                       \textbf{0.912} &  &
                       \textbf{0.881} & 
                       \textbf{0.913} & 
                       \textbf{0.887} & 
                       \textbf{0.897}  \\ 
\hline
\multirow{7}{*}{20\%}  & Transformer                                  & 0.631                  & 0.711                  & 0.622                  & 0.632                  &  & 0.623                  & 0.659                  & 0.614                  & 0.618                   \\
                       & Trans-mean                                   & 0.612                  & 0.742                  & 0.635                  & 0.641                  &  & 0.568                  & 0.594                  & 0.532                  & 0.553                   \\
                       & GRU-D                                        & 0.646                  & 0.733                  & 0.635                  & 0.648                  &  & 0.648                  & 0.698                  & 0.658                  & 0.672                   \\
                       & SeFT                                         & 0.357                  & 0.421                  & 0.381                  & 0.350                  &  & 0.342                  & 0.349                  & 0.346                  & 0.333                   \\
                       & mTAND                                        & 0.332                  & 0.369                  & 0.377                  & 0.373                  &  & 0.456                  & 0.492                  & 0.490                  & 0.490                   \\
                       & RAINDROP                                     & 0.665                  & 0.720                  & 0.679                  & 0.651                  &  & 0.713                  & 0.758                  & 0.725                  & 0.734                   \\
                       & Ours                                         & \textbf{0.863}& 
                       \textbf{0.886}& 
                       \textbf{0.867}& 
                       \textbf{0.876}&  & 
                       \textbf{0.869}& 
                       \textbf{0.896}& 
                       \textbf{0.878}& 
                       \textbf{0.886}  \\ 
\hline
\multirow{7}{*}{30\%}  & Transformer                                  & 0.316                  & 0.264                  & 0.240                  & 0.190                  &  & 0.520                  & 0.552                  & 0.501                  & 0.484                   \\
                       & Trans-mean                                   & 0.425                  & 0.453                  & 0.370                  & 0.339                  &  & 0.651                  & 0.638                  & 0.679                  & 0.649                   \\
                       & GRU-D                                        & 0.451                  & 0.517                  & 0.421                  & 0.472                  &  & 0.580                  & 0.632                  & 0.582                  & 0.593                   \\
                       & SeFT                                         & 0.327                  & 0.279                  & 0.345                  & 0.280                  &  & 0.317                  & 0.310                  & 0.320                  & 0.280                   \\
                       & mTAND                                        & 0.275                  & 0.312                  & 0.306                  & 0.308                  &  & 0.347                  & 0.434                  & 0.363                  & 0.395                   \\
                       & RAINDROP                                     & 0.524                  & 0.609                  & 0.513                  & 0.484                  &  & 0.603                  & 0.681                  & 0.603                  & 0.619                   \\
                       & Ours                                         & \textbf{0.785}
                       & \textbf{0.839}
                       & \textbf{0.795}
                       & \textbf{0.808}
                       &  & \textbf{0.780}
                       & \textbf{0.840}
                       & \textbf{0.788}
                       & \textbf{0.808}  \\ 
\hline
\multirow{7}{*}{40\%}  & Transformer                                  & 0.230                  & 0.074                  & 0.145                  & 0.069                  &  & 0.438                  & 0.446                  & 0.405                  & 0.402                   \\
                       & Trans-mean                                   & 0.257                  & 0.091                  & 0.185                  & 0.099                  &  & 0.487                  & 0.558                  & 0.542                  & 0.551                   \\
                       & GRU-D                                        & 0.464                  & 0.645                  & 0.426                  & 0.443                  &  & 0.477                  & 0.634                  & 0.445                  & 0.475                   \\
                       & SeFT                                         & 0.263                  & 0.299                  & 0.273                  & 0.223                  &  & 0.268                  & 0.241                  & 0.280                  & 0.233                   \\
                       & mTAND                                        & 0.194                  & 0.151                  & 0.202                  & 0.170                  &  & 0.237                  & 0.339                  & 0.264                  & 0.293                   \\
                       & RAINDROP                                     & 0.525                  & 0.534                  & 0.486                  & 0.447                  &  & 0.570                  & 0.654                  & 0.567                  & 0.589                   \\
                       & Ours                                         & \textbf{0.681}
                       & \textbf{0.784}
                       & \textbf{0.674}
                       & \textbf{0.693}
                       &  & \textbf{0.677}
                       & \textbf{0.766}
                       & \textbf{0.681}
                       & \textbf{0.703}  \\ 
\hline
\multirow{7}{*}{50\%}  & Transformer                                  & 0.214                  & 0.027                  & 0.125                  & 0.044                  &  & 0.432                  & 0.520                  & 0.369                  & 0.419                   \\
                       & Trans-mean                                   & 0.213                  & 0.028                  & 0.125                  & 0.046                  &  & 0.464                  & 0.591                  & 0.431                  & 0.465                   \\
                       & GRU-D                                        & 0.373                  & 0.296                  & 0.328                  & 0.266                  &  & 0.497                  & 0.524                  & 0.425                  & 0.475                   \\
                       & SeFT                                         & 0.247                  & 0.159                  & 0.253                  & 0.182                  &  & 0.264                  & 0.230                  & 0.275                  & 0.235                   \\
                       & mTAND                                        & 0.169                  & 0.126                  & 0.170                  & 0.139                  &  & 0.209                  & 0.351                  & 0.230                  & 0.277                   \\
                       & RAINDROP                                     & 0.466                  & 0.445                  & 0.424                  & 0.380                  &  & 0.472                  & 0.594                  & 0.448                  & 0.476                   \\
                       & Ours                                         & \textbf{0.610}
                       & \textbf{0.726}
                       & \textbf{0.614}
                       & \textbf{0.616}
                       &  & \textbf{0.650}
                       & \textbf{0.754}
                       & \textbf{0.647}
                       & \textbf{0.676}  \\
\hline
\end{tabular}
}
}
\end{table}

To further demonstrate the performance of our proposed model, we compared it with six different methods in multivariate time series classification with missing values. The six methods are: Transformer \cite{vaswani2017attention}, Trans-mean \cite{zhang2021graph}, GRUD \cite{che2018recurrent}, SeFT \cite{pmlr-v119-horn20a}, mTAND \cite{shukla2021multi}, and RAINDROP \cite{zhang2021graph}. As mentioned earlier, we follow the experimental settings in previous work \cite{zhang2021graph} in this section. To be fair, the results of the compared methods are duplicated from the work \cite{zhang2021graph}, and our model is trained and tested on its training and testing data.

The results of the comparison are presented in Table \ref{exp_PAM}. From the results, it can be observed that the proposed method performed the best overall, with the highest accuracy and F1-score for both fixed and random sensor out across the five different missing ratios. The RAINDROP method shows the second-best performance in the five testing cases.

Specifically, when the missing ratio is 10\%, the proposed method achieved an improvement of 16.6\% in accuracy and 21.3\% in F1-score for fixed sensor out and 14.9\% in accuracy and 14.1\% in F1-score for random sensor out when compared to the RAINDROP method. When the missing ratio is 20\%, the proposed method achieved an improvement of 29.8\% in accuracy and 34.6\% in F1-score for fixed sensor out and 21.9\% in accuracy and 20.7\% in F1-score for random sensor out. At a missing ratio of 30\%, the proposed method demonstrated a statistically significant improvement in both accuracy and F1-score when compared to the RAINDROP method, with an increase of 49.8\% and 66.9\% respectively for fixed sensor out, and 29.4\% and 30.5\% respectively for random sensor out. Similarly, at a missing ratio of 40\%, the proposed method exhibited a statistically significant improvement in both accuracy and F1-score when compared to the RAINDROP method, with an increase of 29.7\% and 55.0\% respectively for fixed sensor out, and 18.8\% and 19.4\% respectively for random sensor out. Lastly, at a missing ratio of 50\%, the proposed method again demonstrated a statistically significant improvement in both accuracy and F1-score when compared to the RAINDROP method, with an increase of 30.9\% and 62.1\% respectively for fixed sensor out, and 37.7\% and 42.0\% respectively for random sensor out.

In summary, the proposed method demonstrates superior performance to the RAINDROP method and other methods in the comparison, especially at high missing ratios, indicating that it is a robust and effective approach for handling missing data in multivariate time series classification tasks.

\subsection{Results on Real-world Dataset}

\begin{table}
\renewcommand\arraystretch{1.2}
\centering
\caption{Classification accuracy of 4 real-world incomplete time series data sets.}
\resizebox{\linewidth}{!}{ 
\fontsize{30}{32}\selectfont
\begin{tabular}{c|c|c|c|c|c} 
\hline
\diagbox{Methods}{Dataset} & \begin{tabular}[c]{@{}c@{}}Dodger\\LpDay\end{tabular} & \begin{tabular}[c]{@{}c@{}}Dodger\\LpGame\end{tabular} & \begin{tabular}[c]{@{}c@{}}Dodger\\LpWeek\end{tabular} & \begin{tabular}[c]{@{}c@{}}Melbourne\\Pedestrian\end{tabular}  & Average\\ 
\hline
Zero                    & 0.587                                                 & 0.855                                                  & 0.964                                                  & 0.849                 & 0.814                                         \\ 

BRITS                     & 0.550                                                 & 0.710                                                  & 0.891                                                  & 0.687                  & 0.710                                        \\ 

GRUD                     & 0.363                                                 & 0.804                                                  & 0.717                                                  & 0.935   & 0.705                                                        \\ 

Raindrop                     & 0.375                                                 & 0.478                                                  & 0.978                                                  & 0.640       & 0.618                                                    \\ 

AJRNN  & 0.650   & 0.877  & 0.986 & 0.884     & 0.849                                                     \\ 
\hline
Ours (Bi-GRU)  & 0.625   & 0.855  & 0.971 & 0.930     & 0.845                                                     \\ 
Ours                        
& \textbf{0.755}                               
& \textbf{0.913}                                
& \textbf{0.986}                                
& \textbf{0.958}                  
& \textbf{0.903}                       \\
\hline
\end{tabular}}
\label{real-world}
\vspace{-3mm}
\end{table}

This section presents an evaluation of six different methods for classifying time series data from the four real-world datasets,  and the results are presented in Table \ref{real-world}. The `Zero' method means that the missing values are filled with 0, and then our proposed multi-scale feature learning module is utilized for feature learning and classification. The `B-GRU' method means that the replacement of the imputation module in our method using the bidirectional GRU model.

Our method achieves the highest classification accuracy on all four datasets, with an accuracy of 0.755 on the Dodger LpDay dataset, 0.913 on the Dodger LpGame dataset, 0.986 on the Dodger LpWeek dataset, and 0.958 on the Melbourne Pedestrian dataset. The method performed significantly better than the other methods evaluated in this section, with the closest competitor, the AJRNN method, achieving an accuracy of 0.650 on the Dodger LpDay dataset, 0.877 on the Dodger LpGame dataset, 0.986 on the Dodger LpWeek dataset, and 0.884 on the Melbourne Pedestrian dataset. More specifically, there is an improvement of 6.4\%, 6.8\%, 10.9\%, 27.2\%, 28.1\%, and 46.1\% over AJRNN, B-GRU, Zero, BRITS, GRUD, and Raindrop, respectively, in terms of average accuracy. 

The results of this section suggest that our method is a promising approach for classifying real-world time series data on the four missing value datasets evaluated. It is also demonstrated that bidirectional GRU does not bring a positive effect to the ITSC task.

\subsection{Performance of Multi-scale feature Learning Module}
To further illustrate the performance of the proposed multi-scale feature learning module, we conduct the comparison. The setting for this comparison is that all of the missing values are replaced by zeros, and then different feature learners are applied. Six state-of-the-art methods: RNNFCN \cite{2016Time}, GRUFCN \cite{2018Deep}, MLSTMFCN \cite{2018Multivariate}, InceptionTime \cite{2020InceptionTime}, OmniScale \cite{onmi_tang}, and XCM  \cite{math9233137} are utilized. Due to the page limit, we present the average value of the 68 UCR datasets with four different missing ratios, as shown in Table \ref{TB2}.

According to the results presented in Table \ref{TB2}, it can be concluded that the proposed MSFL method provides a significant improvement in the classification accuracy of incomplete time series data compared to the other models. Specifically, MSFL achieved an average accuracy of 0.664, which is higher than the average accuracy of the other models, which ranges between 0.589 and 0.649. 

When looking at the relative improvement, our proposed MSFL improved the average accuracy by 12.2\% compared to RNNFCN, 11.2\% compared to GRUFCN, 11.8\% compared to MLSTMFCN, 2.46\% compared to InceptionTime, 2.31\% compared to OmniScale, and 12.7\% compared to XCM. Another interesting finding is that the MLSTMFCN achieves the same average accuracy with the XCM. It is also worth noting that the improvement in performance is more substantial when the missing ratio is higher.

Overall, the results of this comparison indicate that the MSFL method is a superior method for classifying incomplete time series data compared to the other state-of-the-art models.

\begin{table}
\renewcommand\arraystretch{1.2}
\centering
\caption{Average classification accuracy of different time series classification models after performing zero imputation on UCR univariate time series datasets.}
\resizebox{\linewidth}{!}{ 
\fontsize{10}{12}\selectfont
\begin{tabular}{c|c|c|c|c|c} 
\hline
\diagbox{Methods}{Ratio} & 20\%                                          & 40\%                      & 60\%                      & 80\%    & Average                    \\ 
\hline
RNNFCN                           & 0.664                                       & 0.617                  & 0.575                   & 0.512  & 0.592                 \\
GRUFCN                           & 0.674                                       & 0.630                   & 0.576                  & 0.509     & 0.597              \\
MLSTMFCN                         & 0.674                                       & 0.625                   & 0.569                  & 0.507       & 0.594            \\
InceptionTime                    & 0.725                             & 0.682         & 0.633           & 0.554     & 0.648       \\
OmniScale                        & 0.721                               & 0.681          & 0.635           & 0.560       & 0.649     \\
XCM                              & 0.678                                       & 0.621                   & 0.567                   & 0.489          &    0.589       \\ 
\hline
\textbf{MSFL (Ours)}   & \textbf{0.750} & \textbf{0.699} & \textbf{0.641} & \textbf{0.565} & \textbf{0.664}  \\ \hline
\end{tabular}
}
\label{TB2}
\vspace{-8pt}
\end{table}

\begin{figure}[htbp]
 \vspace{-3mm}
	\centering
			\includegraphics[width=\linewidth]{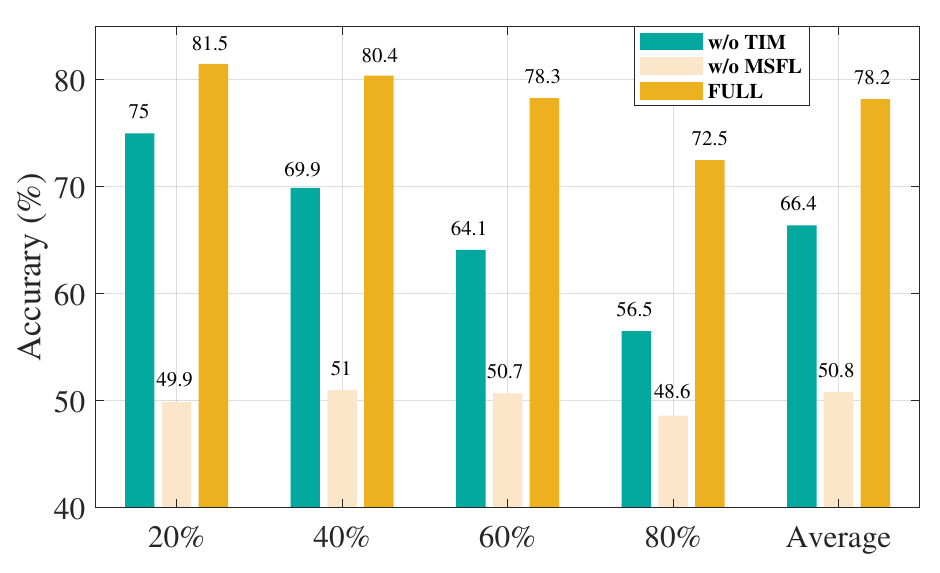}
	\caption{Ablation analysis.}
	\label{ablation}
\end{figure}

\subsection{Ablation and Sensitivity Analysis}

To further investigate the importance of each module in the proposed model, we perform the ablation analysis. To achieve this goal, two different variants are generated: \textbf{1) w/o TIM}: The temporal imputation module is removed and all of the missing values are filled with zeros. \textbf{2) w/o MSFL}: The multi-scale feature learning module is not used and a linear transformation is employed as the classifier as the previous work did \cite{ma_ajrnn}. \textbf{3) Full}: all modules are utilized. The result is presented in Fig.~\ref{ablation}. 

The ablation is performed on the UCR univariate time series datasets. The average performance across various missing ratios and the average performance over the four missing ratios are utilized for comparisons. From the result, we can know that most model drop happens when the MSFL is removed. When the TIM is not used, there is also a significant performance decrease. We can conclude that there highest performance can be obtained by using both modules.

In the sensitivity analysis, the focus is on investigating several key parameters and how they affect the performance of the model. Four main parameters were considered in this study: the number of hidden nodes in the TIM, the impact of varying kernel sizes of the MSFL, the number of layers in the MSFL model, and the effect of the two hyper-parameters in the loss function on the model's performance.

\begin{figure}
     \centering
     \begin{subfigure}[b]{0.225\textwidth}
         \centering
         \includegraphics[width=\textwidth]{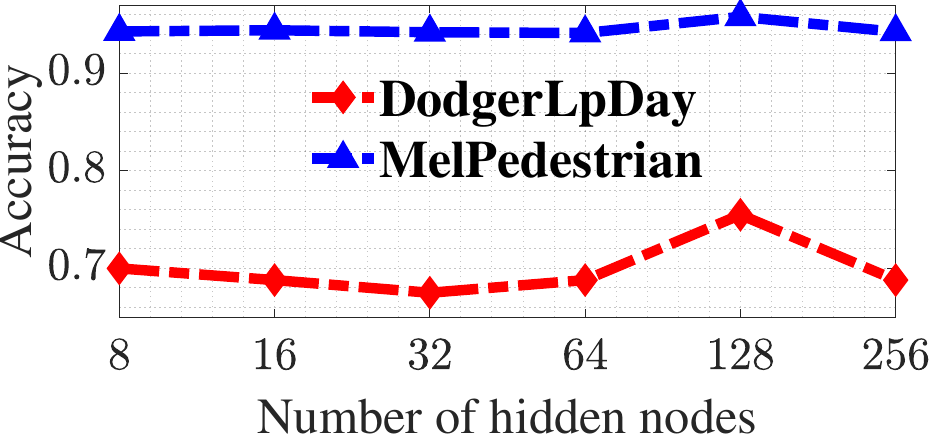}
         \caption{SA for TIM}
     \end{subfigure}
     \hfill
     \begin{subfigure}[b]{0.225\textwidth}
         \centering
    \includegraphics[width=\textwidth]{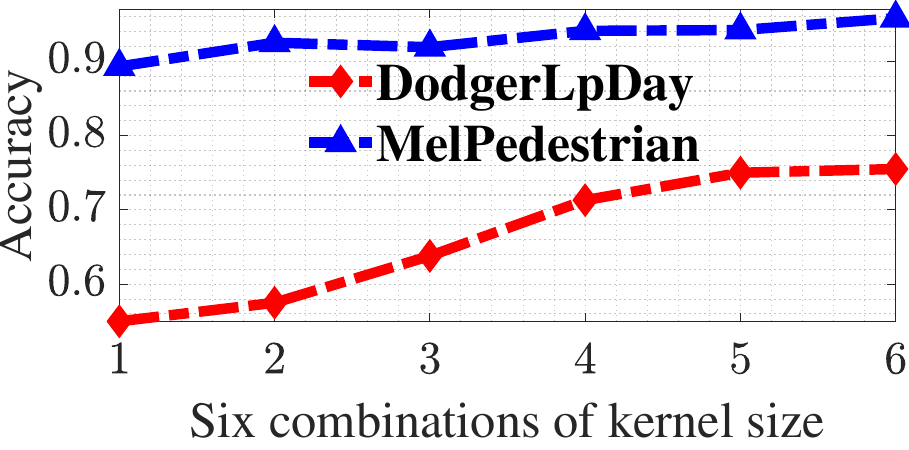}
         \caption{SA for kernel size in MSFL}
     \end{subfigure}

     \begin{subfigure}[b]{0.225\textwidth}
         \centering
         \includegraphics[width=\textwidth]{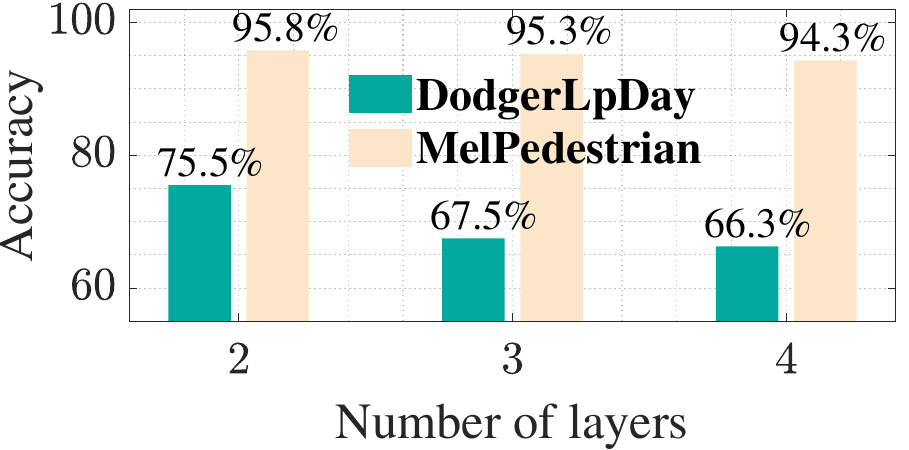}
         \caption{SA for layers in MSFL}
     \end{subfigure}
     \hfill
     \begin{subfigure}[b]{0.225\textwidth}
         \centering
         \includegraphics[width=\textwidth]{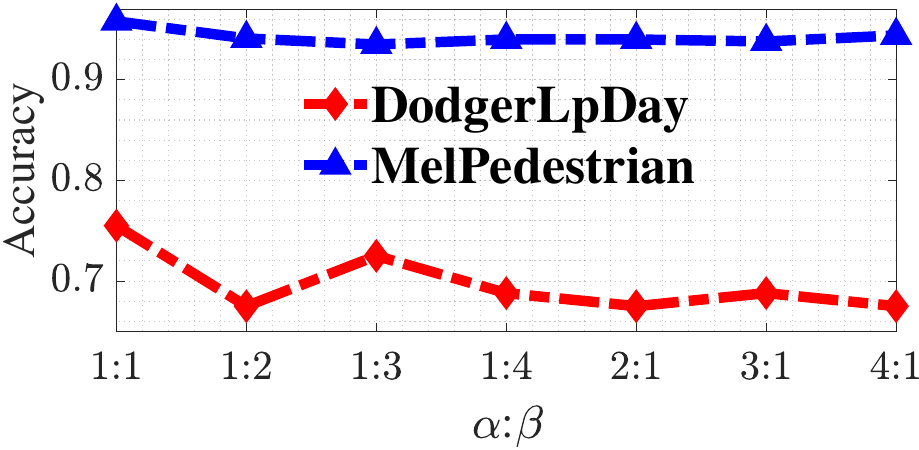}
         \caption{SA for loss function}
     \end{subfigure}     
        \caption{Sensitivity analysis.}
        \label{fig:sa}
        \vspace{-6mm}
\end{figure}

The results of the sensitivity analysis are shown in Fig.~\ref{fig:sa}. When analyzing the number of hidden nodes in the TIM, it is found that the highest performance is achieved when the number of hidden nodes is 128. In terms of the number of layers in the MSFL model, the highest performance is achieved when there are 2 layers. The ratio of $\alpha$ and $\beta$ in the loss function was also examined, and it is found that the best performance is achieved when the ratio is 1:1. Finally, the impact of varying kernel sizes of the MSFL is analyzed, and there are six different combinations of kernel size: `7', `7, 11', `7, 11, 15', `7, 11,15,19', `7, 11,15,19,23', `7, 11,15,19,23,27'. It is found that the higher the number of kernels, the higher the performance of the model. In particular, the highest performance is achieved when the kernel sizes were 7, 11, 15, 19, 23, and 27.

\subsection{Model Interpretation}

\begin{figure}
     \centering
     \begin{subfigure}[b]{0.225\textwidth}
         \centering
         \includegraphics[width=\textwidth]{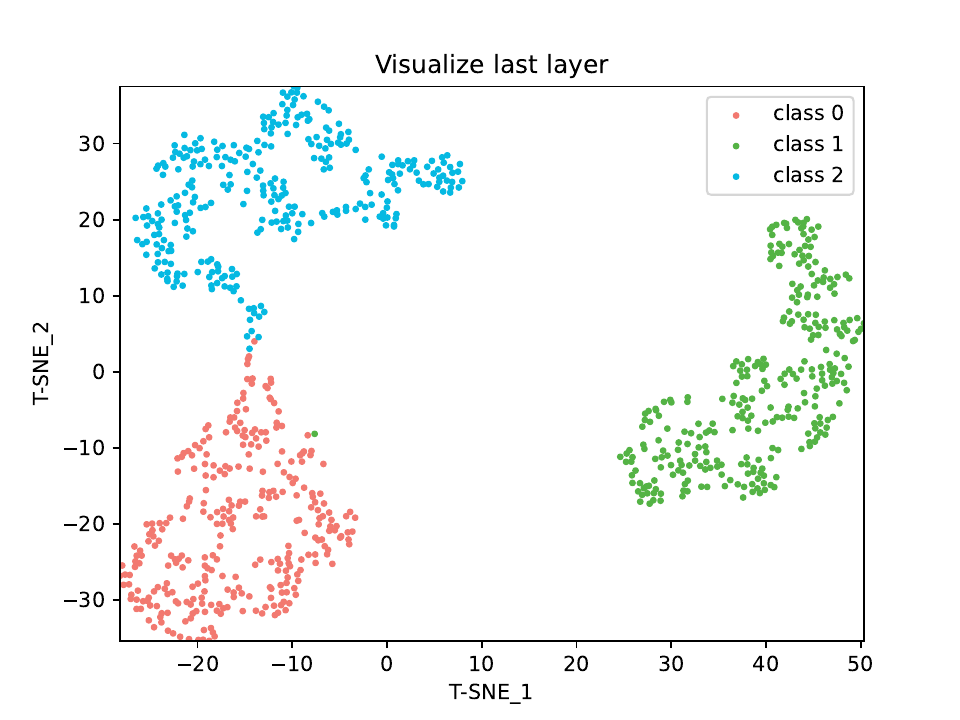}
         \caption{20\%}
     \end{subfigure}
     \hfill
     \begin{subfigure}[b]{0.225\textwidth}
         \centering
    \includegraphics[width=\textwidth]{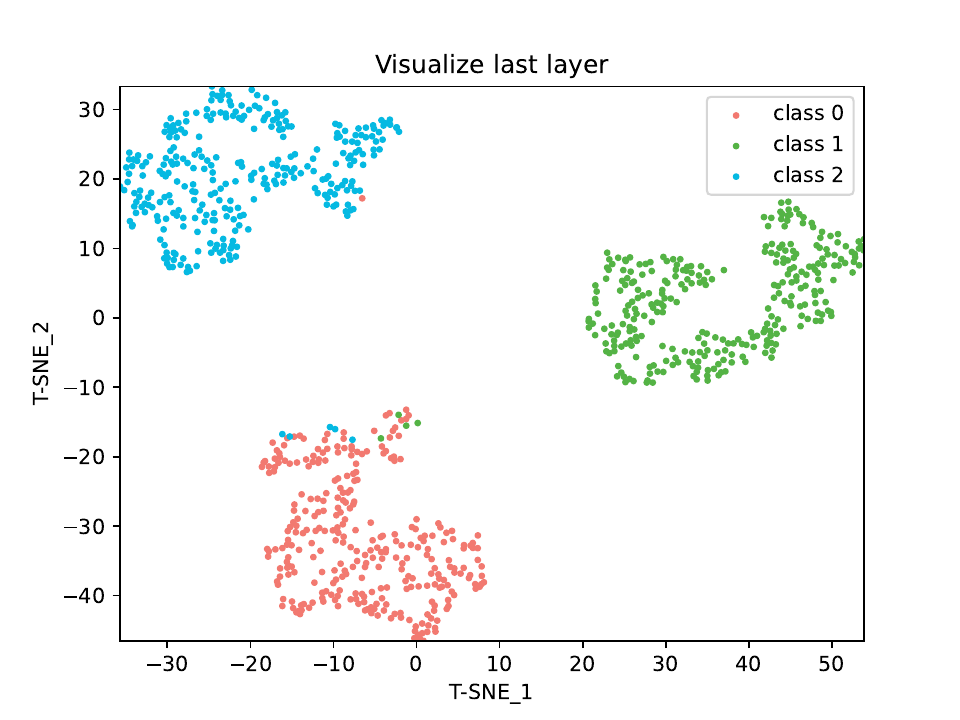}
         \caption{40\%}
     \end{subfigure}

     \begin{subfigure}[b]{0.225\textwidth}
         \centering
         \includegraphics[width=\textwidth]{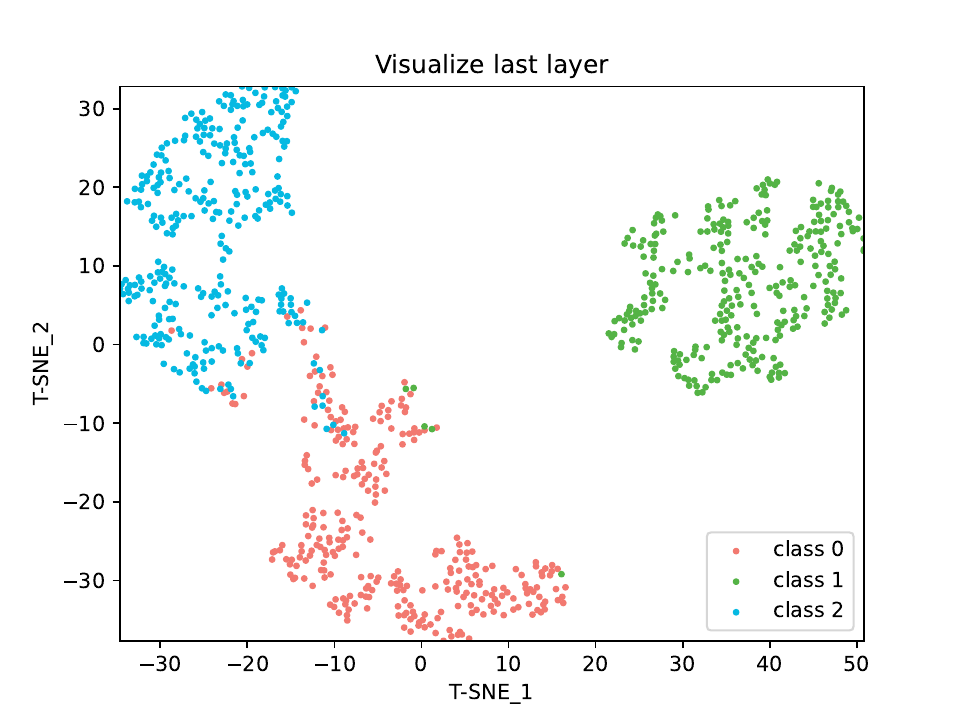}
         \caption{60\%}
     \end{subfigure}
     \hfill
     \begin{subfigure}[b]{0.225\textwidth}
         \centering
         \includegraphics[width=\textwidth]{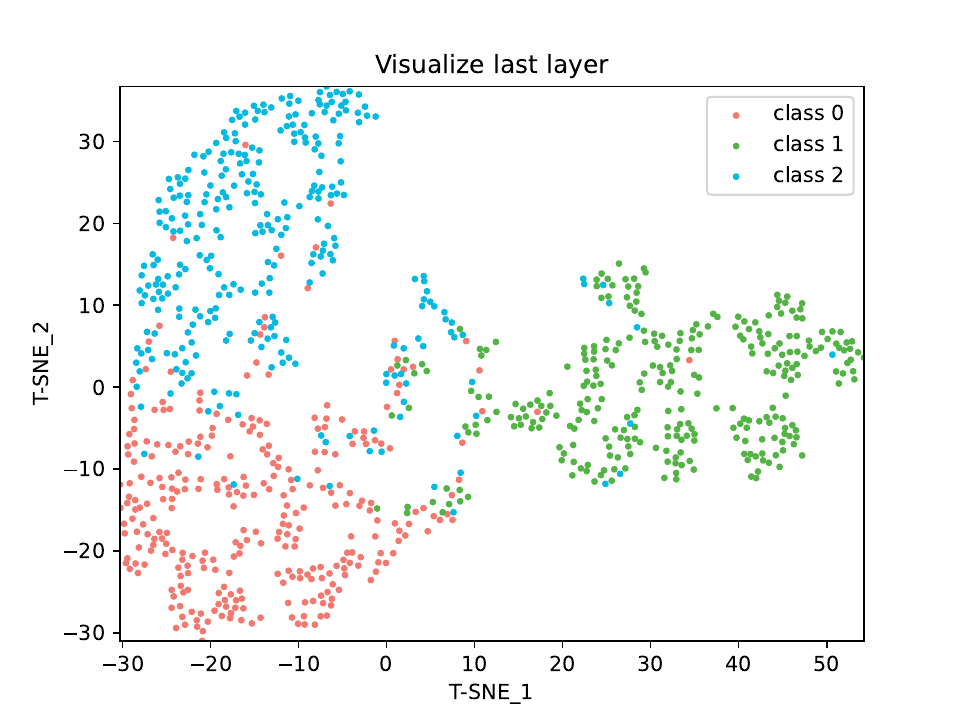}
         \caption{80\%}
     \end{subfigure}     
        \caption{The t-SNE visualization of CBF dataset.}
        \label{fig:Interpretation}
\end{figure}

\begin{figure}
     \centering
     \begin{subfigure}[b]{0.225\textwidth}
         \centering
         \includegraphics[width=\textwidth]{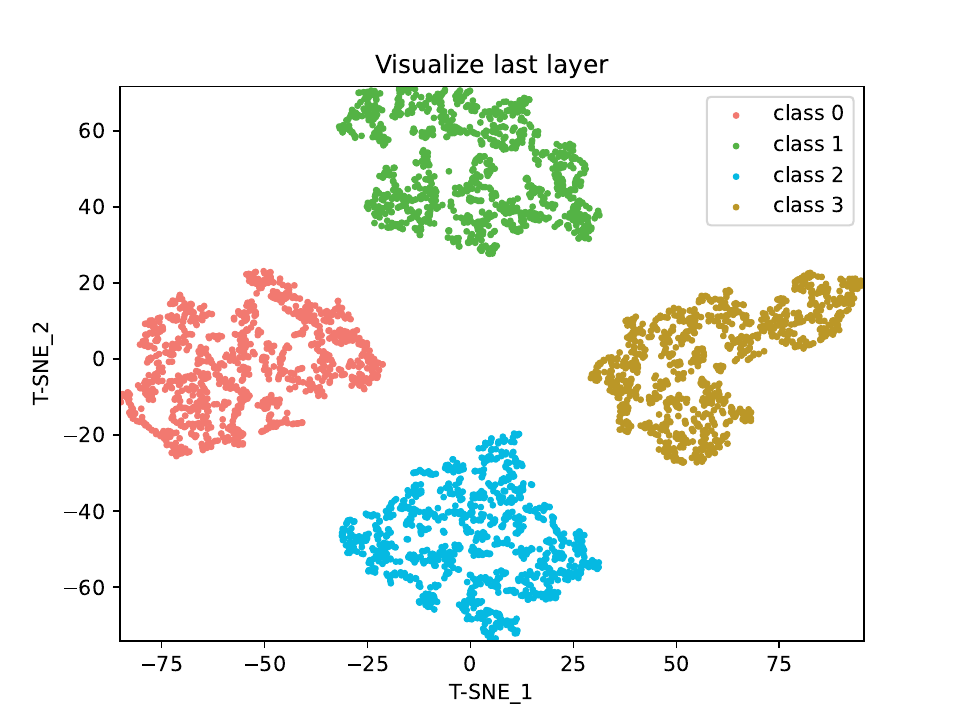}
         \caption{20\%}
     \end{subfigure}
     \hfill
     \begin{subfigure}[b]{0.225\textwidth}
         \centering
    \includegraphics[width=\textwidth]{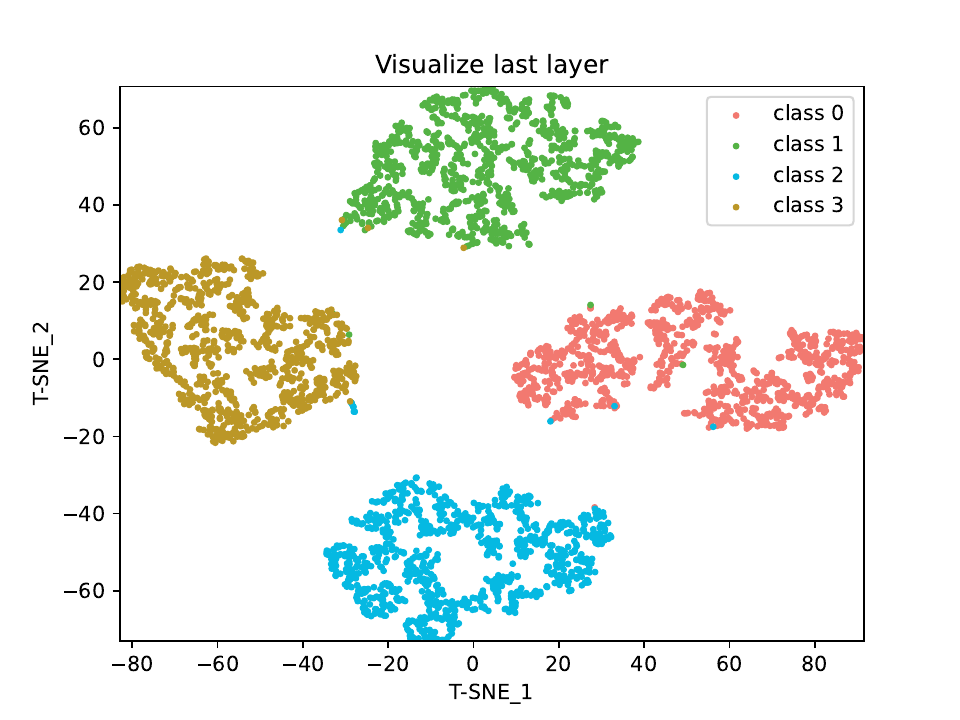}
         \caption{40\%}
     \end{subfigure}

     \begin{subfigure}[b]{0.225\textwidth}
         \centering
         \includegraphics[width=\textwidth]{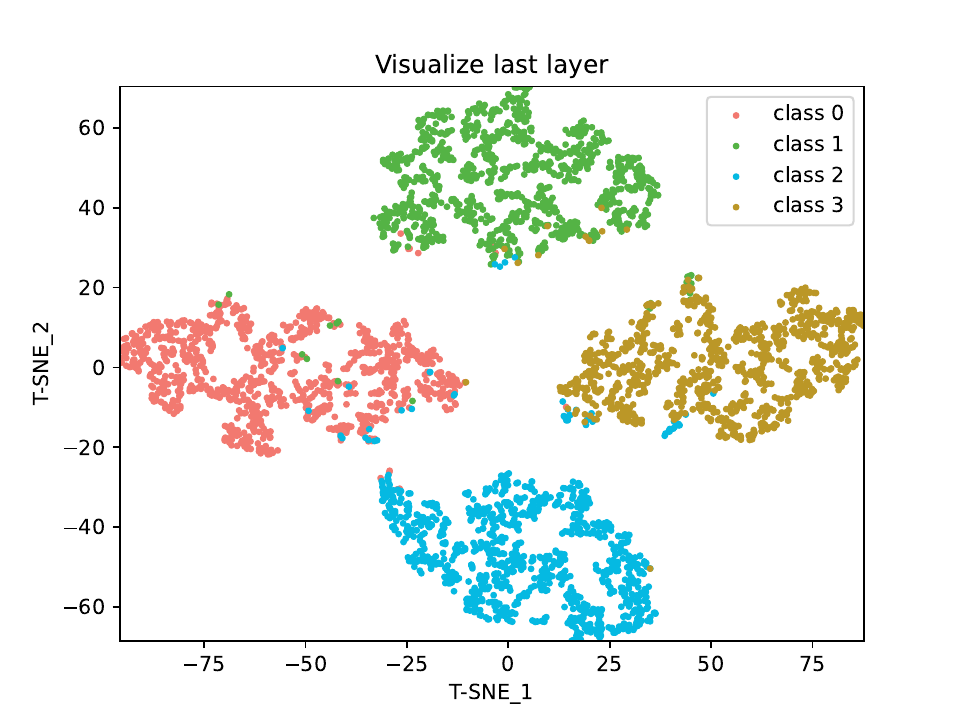}
         \caption{60\%}
     \end{subfigure}
     \hfill
     \begin{subfigure}[b]{0.225\textwidth}
         \centering
         \includegraphics[width=\textwidth]{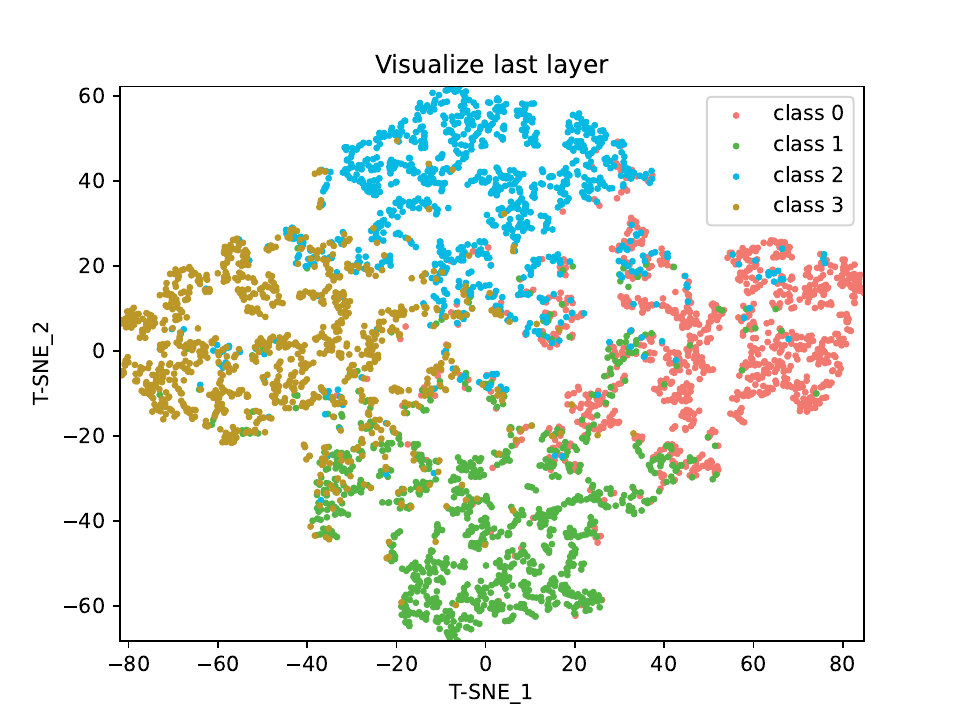}
         \caption{80\%}
     \end{subfigure}     
        \caption{The t-SNE visualization of the TwoPattern dataset.}
        \label{fig:Interpretation2}
     \vspace{-3mm}
\end{figure}

In order to gain a deeper understanding of the mechanisms underlying the proposed model, the features in the last feature layer, which serves as the input to the classifier, were analyzed. Two datasets, CBF (with three classes) and TwoPattern (with four classes), are selected as examples to investigate the robustness of the method as the missing data ratio increases. To facilitate visualization, t-SNE is utilized to map the learned features into a 2D space.

The results of this analysis are illustrated in Fig.~\ref{fig:Interpretation} and Fig.~\ref{fig:Interpretation2}. When the missing ratio is between 20\% and 60\%, it can be observed that the features extracted by the proposed model form well-separated clusters, indicating a high level of performance. However, as the missing ratio increases, the performance gradually decreases. At a missing ratio of 80 percent, the features learned appear to be disorganized, likely due to the accumulation of errors resulting from inaccurate imputed values. 

\section{Conclusion}\label{conclusion}
In this study, we present a novel joint learning framework for addressing classification tasks involving incomplete time series data, which combines imputation and classification training. The proposed model was evaluated on 68 univariate time series datasets from the UCR archive, as well as a multivariate time series dataset with varying missing data ratios. Additionally, the model performance was tested on four real-world datasets with missing information. The experimental results demonstrate that our model outperforms state-of-the-art approaches for incomplete time series classification, particularly in scenarios with high levels of missing data. Additionally, the results demonstrate the superiority of our proposed multi-scale feature learning module in extracting features from incomplete time series data compared to other state-of-the-art architectures. Our emphasis on the classification task enables the network to learn from both imputed and observed values, optimizing imputation for improved classification performance.

\bibliographystyle{IEEEtran}
\bibliography{sample}












\end{document}